\documentclass[twocolumn]{svjour3}          
\smartqed  
\usepackage{natbib}

\usepackage{multicol}
\usepackage{xcolor}
\usepackage[bookmarks=true]{hyperref}

\definecolor{lightblue}{rgb}{0,0.2,1}
\definecolor{black}{rgb}{0,0,0}
\hypersetup{
    colorlinks=true,%
    citebordercolor=white,%
    filebordercolor=white,%
    linkbordercolor=white,%
    urlbordercolor=white,%
    citecolor=lightblue,%
    linkcolor=lightblue,%
    urlcolor=lightblue%
}

\usepackage{amsmath,amssymb,mathrsfs}
\usepackage{mathtools}
\usepackage{algorithm}
\usepackage[noend]{algorithmic}
\usepackage[Symbol]{upgreek}
\usepackage{comment}
\usepackage{graphicx}
\usepackage{graphics}
\usepackage[tight]{subfigure}
\usepackage{float}
\usepackage{multirow}
\usepackage{array}
\usepackage{enumerate}
\usepackage{sidecap}
\usepackage[all]{xy}
\usepackage{authblk}
\usepackage{tikz}
\usepackage{tabularx,pbox,booktabs,caption}

\newcommand*\circled[1]{\tikz[baseline=(char.base)]{
            \node[shape=circle,fill,inner sep=1pt] (char) {\textcolor{white}{#1}};}}
            
\graphicspath{{./figs/}}

\numberwithin{algorithm}{section}  

\newcounter{tecounter}
\begin{document}

\title{Pseudo-Trilateral Adversarial Training for Domain Adaptive Traversability Prediction
}

\author{Zheng Chen        \and
       Durgakant Pushp  \and
       Jason M. Gregory \and
       Lantao Liu
}


\institute{Zheng Chen, Durgakant Pushp, Lantao Liu are with Luddy School of Informatics, Computing, and Engineering, Indiana University, Bloomington, IN 47408, USA\\
              \email{\{zc11, dpushp, lantao\}@iu.edu}\\
Jason M. Gregory is with U.S. Army Research Laboratory\\
\email{jason.m.gregory1.civ@army.mil}\\
Corresponding author: Lantao Liu 
}

\date{Received: date / Accepted: date}

\maketitle

\begin{abstract}

\keywords{Domain adaptation \and Semantic segmentation \and Adversarial training \and Class alignment}

\end{abstract}
Traversability prediction is a fundamental perception capability for autonomous navigation. Deep neural networks (DNNs) have been widely used to predict traversa-bility during the last decade. The performance of DNNs is significantly boosted by exploiting a large amount of data. However, the diversity of data in different domains imposes significant gaps in the prediction performance. In this work, we make efforts to reduce the gaps by proposing a novel pseudo-trilateral adversarial model that adopts a coarse-to-fine alignment (CALI) to perform \textit{unsupervised domain adaptation} (UDA). Our aim is to transfer the perception model with high data efficiency, eliminate the prohibitively expensive data labeling, and improve the generalization capability during the adaptation from easy-to-access \textit{source domains} to various challenging \textit{target domains}. Existing UDA methods usually adopt a bilateral zero-sum game structure. We prove that our CALI model --- a pseudo-trilateral game structure is advantageous over existing bilateral game structures. This proposed work bridges theoretical analyses and algorithm designs, leading to an efficient UDA model with easy and stable training. We further develop a variant of CALI --- Informed CALI (ICALI), which is inspired by the recent success of mixup data augmentation techniques and mixes informative regions based on the results of CALI. This mixture step provides an explicit bridging between the two domains and exposes under-performing classes more during training. We show the superiorities of our proposed models over multiple baselines in several challenging domain adaptation setups. To further validate the effectiveness of our proposed models, we then combine our perception model with a visual planner to build a navigation system and show the high reliability of our model in complex natural environments. 

\begin{figure}
{
\centering
  {\includegraphics[width=0.98\linewidth]{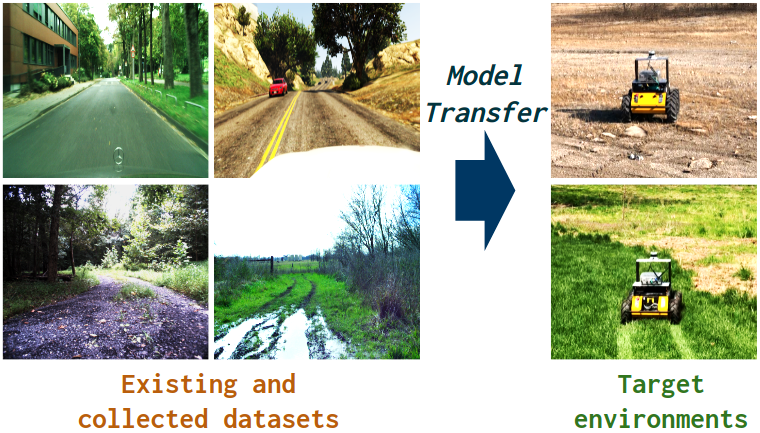} 
  } 
\caption{\small Transferring models from the available domain to the target domain. The existing available data might be from either a simulator or collecting data in certain environments, at a certain time, and with certain sensors. In contrast, the target deployment might have significantly varying environments, time, and sensors.}  \vspace{-20pt}
\label{fig:intro} 
}
\end{figure}

\section{Introduction}
\label{sec:introduction}
We consider the deployment of autonomous robots in real-world unstructured field environments, where the environments can be extremely complex involving random obstacles (e.g., big rocks, tree stumps, man-made objects), cross-domain terrains (e.g., combinations of gravel, sand, wet, uneven surfaces), as well as dense vegetation (tall and low grasses, shrubs, trees). Whenever a robot is deployed in such an environment, it needs to understand which area of the captured scene is navigable. A typical solution to this problem is the visual traversability prediction that can be achieved by learning the {\em scene semantic segmentation} \citep{yang2023learning, jin2021memory}. 

\begin{figure*}
{\centering
  {\includegraphics[width=0.95\linewidth]{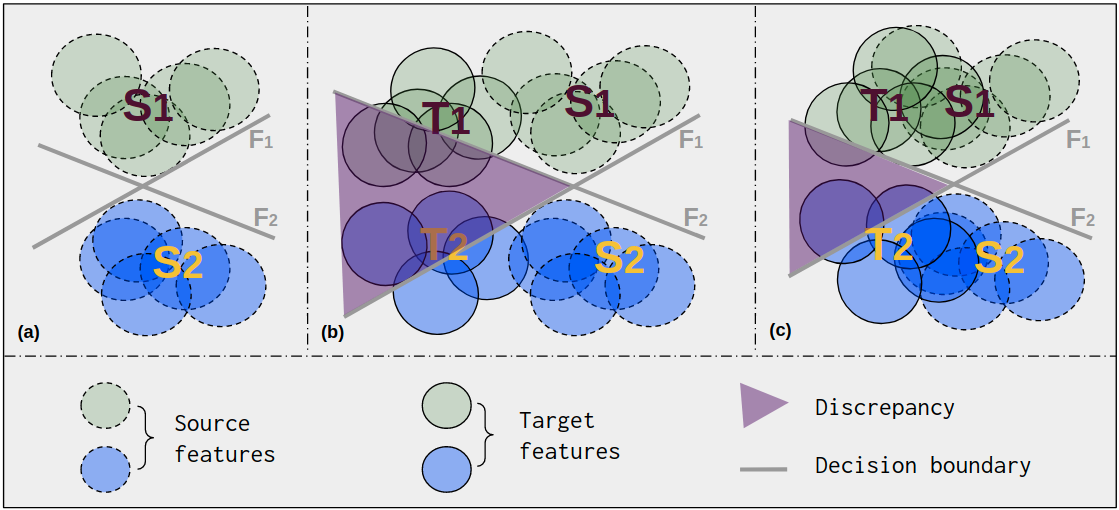} 
  } 
\caption{\small An example to explain the intuition of our CALI model. $S_1/T_1$: Features of class \#1 from source/target data; $S_2/T_2$: Features of class \#2 from source/target data. (a) We first have two well-trained classifiers $F_1$ and $F_2$ on the source domain; (b) The discrepancy can be large for the two trained classifiers on the target domain due to the large domain shift; (c) Our CALI model aims to apply domain alignment to reduce the general domain distance between the source domain and the target domain before performing the class alignment. In this case, the adversarial objective for class alignment --- the discrepancy between the two classifiers, can be reduced to a proper level (see the shadow area change from (b) to (c)). With a well-regularized adversarial objective, the training of CA can be stabilized and the performance can be improved. }
\label{fig:intuition} 
}
\end{figure*}

Visual traversability prediction has been tackled using deep neural networks where the models are typically trained offline with well-labeled datasets. However, there might exist a gap between the data used to train the model and the data when testing. There are several existing datasets for semantic segmentation, e.g., GTA5~\citep{richter2016playing}, SYNTHIA~\citep{Ros_2016_CVPR}, Cityscapes~\citep{Cordts2016Cityscapes}, ACDC~\citep{SDV21}, Dark Zurich~\citep{SDV19}, RUGD\\\citep{RUGD2019IROS}, RELLIS~\citep{jiang2020rellis3d}, and ORFD~\citep{min2022orfd}. Nevertheless, it is usually challenging for existing datasets to well approximate the true distributions of unseen {\em target} environments where the robot is deployed. Even the gradual collection and addition of new training data on an ongoing basis cannot ensure a comprehensive representation of target environments within the distribution. In addition, manually annotating labels for dense predictions, e.g., semantic segmentation, is prohibitively expensive due to the large volumes of data. Therefore, developing a generalization-aware deep model is crucial for 
the robustness, trustworthiness, and safety of
robotic systems considering the demands of the practical deployment of deep perception models and the costs/limits of collecting new data in many robotic applications, e.g., autonomous driving, search and rescue, and environmental monitoring.

To tackle this challenge, a broadly studied framework is \textit{transfer learning}~\citep{pan2009survey} which aims to transfer models between two domains -- \textit{source domain} and \textit{target domain} -- that have related but different data distributions. The prediction on target domain can be considered as a strong generalization since testing data (in target domain) might fall out of the {independently and identically distributed (i.i.d.)}  assumption and follow a very different distribution than the training data (in source domain). The ``transfer"  process has significant meaning to our model development since we can view the available public datasets~\citep{richter2016playing, Cordts2016Cityscapes, RUGD2019IROS, jiang2020rellis3d} as the source domain and treat the data in the to-be-deployed environments as the target domain. In this case, we have access to images and corresponding labels in source domain and images in target domain, but no access to labels in target domain. Transferring models, in this set-up, is called  \textit{Unsupervised Domain Adaptation}~(UDA)~\citep{wilson2020survey, zhang2021survey}.

Domain Alignment (DA)~\citep{ganin2016domain, hoffman2016fcns, hoffman2018cycada, tsai2018learning, vu2019advent} and Class Alignment (CA) \citep{saito2018maximum} are two conventional ways to tackle the UDA problem. DA treats the deep features as a whole. It works well for image-level tasks such as image classification, but has issues with pixel-level tasks such as semantic segmentation \citep{saito2018maximum}, as the alignment of whole distributions ignores the class features and might misalign class distributions, even the whole features from the source domain and target domain are already well-aligned. CA is proposed to solve this issue for dense predictions with multiple classes.

It is natural and necessary to use CA to tackle the UDA of semantic segmentation as we need to consider aligning class features. However, CA can be problematic and might fail to outperform the DA for segmentation, and in a worse case, might have unacceptable \textit{negative transfer}, which means the performance with adaptation is even more degraded than that without adaptation. We empirically found that the training of CA can be unstable and easy to diverge, leading to low performance or even failure of training. Typically, CA adopts a network consisting of a feature extractor and two classification heads. The net is first trained on the source domain such that the decision boundaries of the two heads are both able to well classify source features of different classes. Then the net is trained on the target domain in an adversarial manner, where the discrepancy of the two heads is used as the adversarial objective. During adversarial training, the goal of the feature extractor is to generate features such that the discrepancy between the two heads is minimized; while the goal of the classification heads is to adjust the decision boundaries such that the discrepancy is maximized. By doing so the target features are enforced to be aligned with the trained source features. However, adversarial training can be highly unstable especially when the adversarial objective is not properly valued. We conjecture the reason for the problem of CA is the value of the adversarial objective --- the discrepancy between the two heads is too large due to the domain shift, making the two classification heads in a near-optimal state at the beginning of training. In this case, the training of the two classification heads can quickly converge, breaking the equilibrium of the zero-sum game, and leading to a failure of the whole training. 

To address the potential shortcomings of CA, our intuition is to apply a DA as a constraint to reduce the general distance between two domains before applying CA. This will also bring a reduction of the discrepancy value between the two classification heads, thus the adversarial training over the discrepancy can be stabilized. An example to explain our intuition is shown in Fig. \ref{fig:intuition}. To achieve this, we investigate the relationship of the upper bounds of the prediction error on the target domain between DA and CA and provide a theoretical analysis of the upper bounds of target prediction error for the two alignments in the UDA setup. Our theoretical analysis justifies the use of DA as a constraint of CA.

This paper presents an extended and revised version of our recent work CALI \citep{Chen-RSS-22}. To further improve CALI, we develop a mixture-based supervision augmented version of CALI --- Informed CALI (ICALI), which introduces an extra supervised learning process following the pseudo-trilateral training of CALI. The data for the extra learning is built by mixing informative regions from the source domain and target domain. The \textit{informative regions} are defined as image areas occupied by under-performing classes during training. The advantages of ICALI are twofold: (a) Mixing the data from different regions leads to new data augmentations and further boosts the representation learning; (b) By mixing data \textit{informatively}, the under-performing classes are exposed to the model more frequently such that the performance of those classes are improved. 

In summary, our contributions include
\begin{itemize}
    \item 
    We prove that with proper assumptions, the upper bound of CA is upper bounded by the upper bound of DA. This indicates that constraining the training of CA using DA can be beneficial. 
    We then propose a novel concept of \textit{pseudo-trilateral game structure}~(PTGS) for integrating DA and CA.
    \item
    We propose an efficient coarse-to-fine alignment bas-ed UDA model, named CALI, for traversability prediction. The new proposal includes a trilateral network structure, novel training losses, and an alternative training process. Our model design is well supported by theoretical analysis. It is also easy and stable to train and converge.
    \item 
    Based on the alignment of CALI, we further propose a new variant named ICALI, which achieves better performance than CALI by introducing new data augmentations and extra supervised training.
    \item
    We show significant advantages of our proposed mo-del compared to several baselines in multiple challenging public datasets and one self-collected dataset. 
    We combine the proposed segmentation model and a visual planner to build a visual navigation system. 
\end{itemize}

\section{Related Work}
\label{sec:related work}
\textbf{Semantic Segmentation:} Semantic segmentation aims to predict a unique human-defined semantic class for each pixel in the given images. With the prosperity of deep neural networks, the performance of semantic segmentation has been boosted significantly, especially by the advent of FCN~\citep{long2015fully} that first proposes to use deep convolutional neural nets to predict segmentation. The following works try to improve the FCN performance by multiple proposals, e.g., using different sizes of kernels or dilation rates to aggregate multi-scale features \citep{chen2017deeplab, chen2017rethinking, yu2015multi}; building image pyramids to create multi-resolution inputs~\citep{zhao2017pyramid}; applying probabilistic graph to smooth the prediction~\citep{liu2017deep}; and compensating features in deeper level by an encoder-decoder structure~\citep{ronneberger2015u}. 

Recently, Transformers \citep{vaswani2017attention, dosovitskiy2020image} have gained huge popularity for various vision tasks including semantic segmentation. Transformer-based models employ an attention mechanism to capture the long-range dependencies among pixels. Different Transformer-based segmentation models have been developed, e.g., finer-grained and more globally coherent predictions are achieved for dense predictions by assembling tokens from various stages of the vision transformer into image-like representations at various resolutions~\citep{ranftl2021vision}. Semantic segmentation is treated as a sequence-to-sequence prediction task and a pure transformer without convolution and
resolution reduction are used to encode an image as a sequence of
patches \citep{zheng2021rethinking}. A novel hierarchically structured Transformer encoder is combined with a lightweight MLP decoder to build a simple, efficient yet powerful semantic segmentation
framework \citep{xie2021segformer}. Recent state-of-the-art works \citep{ranftl2021vision, zheng2021rethinking, xie2021segformer} for semantic segmentation heavily rely on Transformer structure.
However, all of those methods belong to fully-supervised learning and the performance might  be degraded catastrophically when a domain shift exists between the training data and the data when deploying.

\textbf{Unsupervised Domain Adaptation:} The main approaches to tackle UDA include adversarial training (a.k.a., \textit{distribution alignment})~\citep{ganin2016domain, hoffman2016fcns, hoffman2018cycada, tsai2018learning, saito2018maximum, vu2019advent, luo2019taking, wang2020classes} and self-training~\citep{zou2018unsupervised, zhang2017curriculum, mei2020instance, hoyer2021daformer}. Self-training maintains a teacher-student framework to conduct the knowledge transfer. The teacher model is trained on the source domain and used to predict segmentation for target images. The predictions from the teacher model are then used as pseudo labels to train the student model. During the training of the student model, existing methods use different ways to identify the erroneous regions, including confidence score \citep{zou2019confidence, zou2018unsupervised} and entropy \citep{xie2022towards, chen2019domain, pan2020unsupervised}. Although self-training is becoming a popular method for segmentation UDA in terms of empirical results, it still lacks a sound theoretical foundation. In this paper, we only focus on the alignment-based methods that not only keep close to the UDA state-of-the-art performance but are also well supported by sound theoretical analyses~\citep{ben2007analysis, blitzer2008learning, ben2010theory}. 

The alignment-based methods adapt models via ali-gning the distributions from the source domain and target domain in an adversarial training process, i.e., making the deep features of source images and target images indistinguishable to a discriminator net. Typical alignment-based approaches to UDA include Domain Alignment \citep{ganin2016domain, hoffman2016fcns, hoffman2018cycada, tsai2018learning, vu2019advent}, which aligns the two domains using global features (aligning the feature tensor from source or target \textit{as a whole}) and Class Alignment \citep{saito2018maximum, luo2019taking, wang2020classes}, which only considers aligning features of each class from source and target, no matter whether the domain distributions are aligned or not. In \citet{saito2018maximum}, the authors are inspired by the theoretical analysis of \citet{ben2010theory} and propose a discrepancy-based model for aligning class features. There is a clear relation between the theory guidance \citep{ben2010theory} and the design of network, loss, and training methods. There are some recent works \citep{luo2019taking, wang2020classes} similar to our proposed work in spirit and show improved results compared to \citet{saito2018maximum}, but it is still unclear to relate the proposed algorithms with theory and to understand why the structure/loss/training is designed as the presented way.

\textbf{Visual Navigation:}
To achieve visual navigation autonomously,  learning-based methods have been widely studied recently \citep{shen2019situational,bansal2020combining}. For example, 
imitation learning based approaches have been largely explored to train a navigation policy that enables a robot to mimic human behaviors or navigate close to certain waypoints without a prior map~\citep{manderson2020vision,hirose2020probabilistic}. To fully utilize the known dynamics model of the robot, a semi-learning-based scheme is also proposed \citep{bansal2020combining} to combine optimal control and deep neural network to navigate through unknown environments. A large amount of work on visual navigation can also be found in the computer vision community, such as \citet{shen2019situational, gupta2017cognitive, chaplot2020neural, gupta2017unifying, wu2019bayesian}, all of which use full-learning-based methods to train navigation policies, which work remarkably well when training data is sufficient but can fail frequently if no or very limited data is available.

\section{Background and Preliminary Materials}
\subsection{Expected Errors}
We consider segmentation tasks where the input space is $\mathcal{X}\subset \mathbb{R}^{H\times W\times 3}$, representing the input RGB images, and the label space is $\mathcal{Y}\subset \left\{ 0, 1\right\}^{H\times W\times K}$, representing the ground-truth $K$-class segmentation images. The label for a single pixel at $(h, w)$ is denoted by a one-hot vector $y^{(h, w)}\in \mathbb{R}^K$ whose elements are by-default 0-valued 
except the $i^{th}$ element is labeled as $1$ if the $i^{th}$ class is specified. Domain adaptation has two domain distributions over $\mathcal{X}\times \mathcal{Y}$, named source domain $\mathcal{D}_s$ and target domain $\mathcal{D}_t$. 
In the setting of UDA for segmentation, we have access to $m_s$ \textit{i.i.d.} samples with labels $\mathcal{U}_s=\left\{\mathbf{x}_{si}, \mathbf{y}_{si} \right\}_{i=1}^{m_s}$ from $\mathcal{D}_s$ and $m_t$ \textit{i.i.d.} samples without labels $\mathcal{U}_t=\left\{\mathbf{x}_{tj} \right\}_{j=1}^{m_t}$ from $\mathcal{D}_t$. 

In the UDA problem, we need to reduce the prediction error on the target domain.  With a slight abuse of notation, we also use $h$ to denote a {\em hypothesis}, which is a function: $\mathcal{X}\rightarrow \mathcal{Y}$. We denote the space of $h$ as $\mathcal{H}$. With the loss function $l(\cdot, \cdot)$, the expected error of $h$ on $\mathcal{D}_s$ is defined as
\begin{equation}
    \label{eq:source_expected_error}
    \epsilon_s(h):=\mathbb{E}_{(x, y)\sim \mathcal{D}_s}l(h(x), y).
\end{equation}
Similarly, we can define the expected error 
of $h$ on $\mathcal{D}_t$ as
\begin{equation}
    \label{eq:target_expected_error}
    \epsilon_t(h):=\mathbb{E}_{(x, y)\sim \mathcal{D}_t}l(h(x), y).
\end{equation}

\subsection{Upper Bounds for Expected Errors}
Two important upper bounds related to the source and target error are given in \citet{ben2010theory}. 

\circled{1} The first upper bound is given in the following theorem. 

\textbf{Theorem 1} 
For a hypothesis $h$, \begin{equation}
    \label{eq:theorem1}
    \epsilon_t(h)\leq \epsilon_s(h)+d_1(\mathcal{D}_s, \mathcal{D}_t)+\lambda,
\end{equation}
where $d_1(\cdot, \cdot)$ is the $L^1$ divergence for two distributions, and the constant term $\lambda$ does not depend on any $h$. 
However, it is claimed in \citet{ben2010theory} that the bound with $L^1$ divergence cannot be accurately estimated from finite samples, and using $L^1$ divergence can unnecessarily inflate the bound. Another divergence measure is thus introduced to replace the $L^1$ divergence with a new bound derived. The new measure is defined as follows,

\textbf{Definition 1} 
Given two domain distributions $\mathcal{D}_s$ and $\mathcal{D}_t$ over $\mathcal{X}$, and a hypothesis space $\mathcal{H}$ that has finite VC dimension, the $\mathcal{H}$-divergence between $\mathcal{D}_s$ and $\mathcal{D}_t$ is defined as
\begin{equation}
    \label{eq:h_divergence}
    \begin{aligned}
        d_{\mathcal{H}}(\mathcal{D}_s, \mathcal{D}_t) = 2\sup_{h\in \mathcal{H}} | &\text{P}_{x\sim \mathcal{D}_s} \left [ h(x)=1 \right ] - \\
        &\text{P}_{x\sim \mathcal{D}_t} \left [ h(x)=1 \right ] |,
    \end{aligned}
\end{equation}
where $\text{P}_{x\sim \mathcal{D}_s}[h(x)=1]$ represents the probability of $x$ belonging to $\mathcal{D}_s$. Same to $\text{P}_{x\sim \mathcal{D}_t} \left [ h(x)=1 \right ]$.

The $\mathcal{H}$-divergence resolves the issues in the $L^1$ divergence. If we replace  $d_1(\mathcal{D}_s, \mathcal{D}_t)$ in Eq.~\eqref{eq:theorem1} with $d_{\mathcal{H}}(\mathcal{D}_s, \mathcal{D}_t)$, then a new upper bound for $\epsilon_t(h)$, named as $\mathbb{UB}_1$,  can be written as
\begin{equation}
    \label{eq:ub1}
    \begin{aligned}
        &\epsilon_t(h)\leq \mathbb{UB}_1,\\
        & \mathbb{UB}_1 = \epsilon_s(h)+d_{\mathcal{H}}(\mathcal{D}_s, \mathcal{D}_t) + \lambda.
    \end{aligned}
\end{equation}

An approach to compute the empirical $\mathcal{H}$-divergence is also proposed in \citet{ben2010theory}, see the below Lemma 1.

\textbf{Lemma 1} 
For a symmetric hypothesis class $\mathcal{H}$ (one where for every $h\in \mathcal{H}$, the inverse hypothesis $1-h$ is also in $\mathcal{H}$) and two sample sets 
\begin{equation}
    \label{eq:two_sets}
    \begin{aligned}
        &\mathcal{U}_s=\left\{ x_i, i=1, \cdots, m_s, x_i\sim \mathcal{D}_s\right\},\\
        &\mathcal{U}_t=\left\{ x_j, j=1, \cdots, m_t, x_j\sim \mathcal{D}_t\right\},
    \end{aligned}
\end{equation}
the approximated empirical $\mathcal{H}$-divergence is computed as:
\begin{equation}
    \label{eq:empirical_h_divergence}\
    \begin{aligned}
        \hat{d}_{\mathcal{H}}(\mathcal{D}_s, \mathcal{D}_t) = 2 \Bigg( \Bigg.  1-\min_{\eta \in \mathcal{H}} \Bigg[ \Bigg. &\frac{1}{m_s} \sum_{i=1}^{m_s} \mathbb{I}[\eta(x_i)=0] +\\
        &\frac{1}{m_t}\sum_{j=1}^{m_t} \mathbb{I}[\eta(x_j)=1] \Bigg. \Bigg] \Bigg. \Bigg),
    \end{aligned}
\end{equation}
where $\mathbb{I}[a]$ is an indicator function which is 1 if $a$ is true, and $0$ otherwise.

\circled{2} The second upper bound is based on a new hypothesis called the symmetric difference hypothesis, see the following definition.

\begin{figure*}
{
\centering
  {\includegraphics[width=0.8\linewidth]{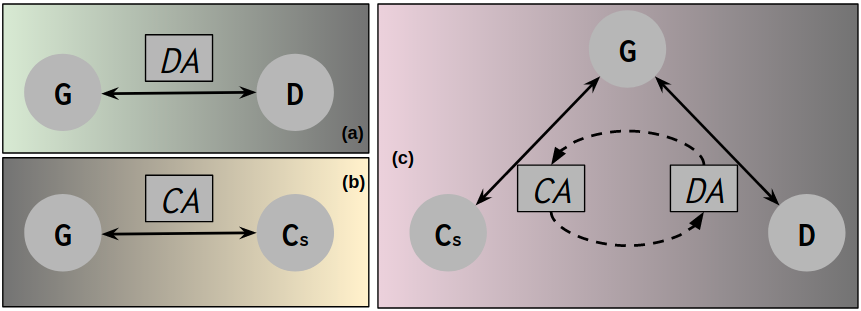} 
  } 
\caption{\small Different game structures. (a) Bilateral game for DA, similar to GANs \citep{goodfellow2014generative}; (b) Bilateral game for CA; (c) Our proposed pseudo-trilateral game structure (PTGS). $G$ is the feature extractor; $D$ is the domain discriminator; $Cs$ are a family of classifiers. }  \vspace{-10pt}
\label{fig:trilateral} 
}
\end{figure*}

\textbf{Definition 2} 
For a hypothesis space $\mathcal{H}$, the symmetric difference hypothesis space $\mathcal{H}\Delta \mathcal{H}$ is the set of hypotheses
\begin{equation}
    \label{eq:hdh_space}
    g\in \mathcal{H}\Delta\mathcal{H} \Leftrightarrow g(x) = h(x)\oplus  h'(x)~~~\text{for some } h, h'\in \mathcal{H},
\end{equation}
where $\oplus$ denotes an XOR operation. Then we can define the $\mathcal{H}\Delta\mathcal{H}$-distance as
\begin{equation}
    \label{eq:hdh_distance}
    \begin{aligned}
    d_{\mathcal{H}\Delta\mathcal{H}}(\mathcal{D}_s, \mathcal{D}_t)=2\sup_{h, h'\in \mathcal{H}} | &\text{P}_{x\sim \mathcal{D}_s} \left [ h(x)\neq h'(x) \right ] - \\
        &\text{P}_{x\sim \mathcal{D}_t} \left [ h(x)\neq h'(x) \right ] |.
    \end{aligned}
\end{equation}
Similar to Eq.~(\ref{eq:ub1}), if we replace $d_1(\mathcal{D}_s, \mathcal{D}_t)$ with the $\mathcal{H}\Delta\mathcal{H}$-distance $d_{\mathcal{H}\Delta\mathcal{H}}(\mathcal{D}_s, \mathcal{D}_t)$,  the second upper bound for $\epsilon_t(h)$, named as $\mathbb{UB}_2$,  can be expressed as
\begin{equation}
    \label{eq:ub2}
    \begin{aligned}
    &\epsilon_t(h)\leq \mathbb{UB}_2,\\
    &\mathbb{UB}_2 = \epsilon_s(h)+d_{\mathcal{H}\Delta\mathcal{H}}(\mathcal{D}_s, \mathcal{D}_t) + \lambda,
    \end{aligned}
\end{equation}
where $\lambda$ is the same term as in Eq.~(\ref{eq:theorem1}).

The two bounds (Eq.~(\ref{eq:ub1}) and Eq.~(\ref{eq:ub2})) for the target domain error are separately given in \citet{ben2010theory}. It has been independently demonstrated that DA corresponds to optimizing over $\mathbb{UB}_1$ \citep{ganin2016domain}, where optimization over the upper bound $\mathbb{UB}_1$ (Eq.~(\ref{eq:ub1}) with the divergence Eq.~(\ref{eq:empirical_h_divergence})) is proved as equivalent to an adversarial learning with Eq.~(\ref{eq:gan}) and a supervised learning with the source data, and that CA corresponds to optimizing over $\mathbb{UB}_2$ \citep{saito2018maximum}, where the $d_{\mathcal{H}\Delta\mathcal{H}}$ is approximated by the discrepancy between two different classifiers.

Training DA is straightforward since we can easily define binary labels for each domain, e.g., we can use 1 as the source domain label and 0 as the target domain label. Adversarial training over the domain labels can achieve domain alignment. For CA, however, it is difficult to implement as we do not have target labels, hence the target class features are completely unknown to us, thus leading naively using adversarial training over each class impossible. The existing way well supported by theory to perform CA \citep{saito2018maximum} is to indirectly align class features by devising two different classifier hypotheses. The two classifiers have to be well trained on the source domain and are able to classify different classes in the source domain with different decision boundaries. Then considering the shift between source and target domain, the trained two classifiers might have disagreements on target domain classes. Note since the two classifiers are already well trained on the source domain, the agreements of the two classifiers represent those features in the target domain that are close to the source domain, while in contrast, the features where disagreements happen indicate that there is a large shift between the source and the target. We use the disagreements to approximate the distance between the source and the target. If we are able to minimize the disagreements of the two classifiers, then features of each class between source and target will be enforced to be well aligned.

\subsection{Adversarial Training}
A standard way to achieve the alignment for deep models is to use the adversarial training method, which is also used in Generative Adversarial Networks (GANs) \citep{goodfellow2014generative}. Therefore we explain the key concepts of adversarial training using the example of GANs.

GAN is proposed to learn the distribution $p_r$ of a set of given data $\left\{ \mathbf{x} \right\}$ in an adversarial manner. The architecture consists of two networks - a generator $G$, and a discriminator $D$. The $G$ is responsible for generating fake data (with distribution $p_g$) from random noises $\mathbf{z}\sim p_{\mathbf{z}}$ to fool the discriminator $D$ that is instead to accurately distinguish between the fake data and the given data. Optimization of a GAN involves a mini-maximization over a joint loss for $G$ and $D$.
\begin{equation}
    \label{eq:gan}
    \begin{aligned}
        &\min_{G}\max_{D} V(G, D)\\
        &V(G, D) = \mathbb{E}_{\mathbf{x}\sim p_r} \log \left [ D(\mathbf{x}) \right ] + \mathbb{E}_{\mathbf{z}\sim p_{\mathbf{z}}}\log \left [ 1-D(G(z)) \right ],
    \end{aligned}
\end{equation}
where we use $1$ as the real label and $0$ as the fake label. Training with Eq.~(\ref{eq:gan}) is a bilateral game where the distribution $p_g$ is aligned with the distribution $p_r$.

\begin{figure*}
{
\centering
  {\includegraphics[width=0.95\linewidth]{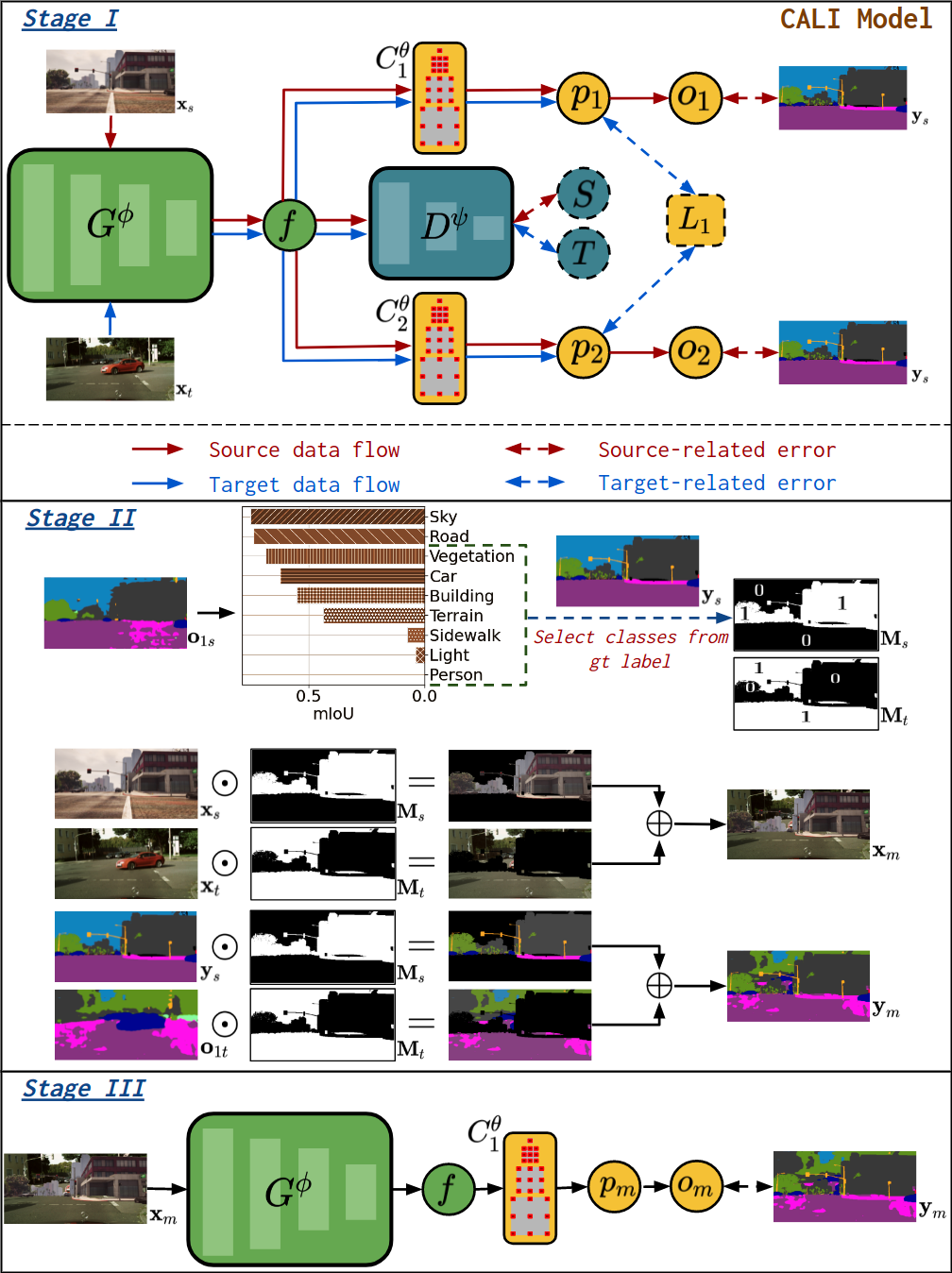} 
  } 
\caption{\small Framework of our proposed model. Stage I: the CALI model. $f$ represents the shared feature map; $p_1/o_1$ and $p_2/o_2$ are the categorical probability/class label predictions; $S/T$ represent domain labels: 1 (source) and 0 (target); $L_1$ represents the $L_1$ distance measure between two vectors. Stage II: Mixing source data and target data. $o_{1s}$ and $o_{1t}$ are label predictions from $C_1^\theta$ for the source image and target image, respectively. Stage III: Extra supervised training using the new mixed data. $p_m/o_m$ are the categorical probability/class label predictions for the mixed data. Stage II \& III make the whole framework as ICALI. See Section 4.2 for more details.}  \vspace{-10pt}
\label{fig:net} 
}
\end{figure*}

\section{Methodology}
\label{sec:methodology}

In this work we investigate the relationship between the $\mathbb{UB}_1$ (Eq. (\ref{eq:ub1})) and $\mathbb{UB}_2$ (Eq. (\ref{eq:ub2})) and prove that $\mathbb{UB}_1$ turns out to be an upper bound of $\mathbb{UB}_2$, meaning DA can be a necessary constraint to CA. This is also consistent with our intuition: DA aligns features globally in a coarse way while CA aligns features locally in a finer way. Constraining CA with DA is actually a coarse-to-fine process. We use the coarse alignment to reduce the general domain distance and regularize the adversarial objective for the fine alignment into a proper level. DA is shown to be a zero-sum game between a feature extractor and a domain discriminator \citep{ganin2016domain}, while CA is proved to be a zero-sum game between a feature extractor and two classifiers \citep{saito2018maximum}. Both DA and CA are bilateral games, see (a) and (b) in Fig. \ref{fig:trilateral}. In this work, we propose a novel concept, \textit{pseudo-trilateral game structure} (PTGS), for efficiently integrating game structures of DA and CA, see (c) in Fig. \ref{fig:trilateral}. Three players are involved in the proposed PTGS, a feature extractor $G$, a domain discriminator $D$, and a family of classifiers $Cs$. The game between $G$ and $Cs$ is the CA while the game between $G$ and $D$ is the DA. According to the identified relation in Eq.~(\ref{eq:bound_relation}), the two upper bounds $\hat{\mathbb{UB}}_1$ and $\hat{\mathbb{UB}}_2$ need to use the same feature, hence we connect the domain alignment and class alignment using a shared feature extractor. Both $D$ and $Cs$ are trying to adjust the $G$ such that the features between source and target generated from $G$ could be well aligned globally and locally. There is no game between the $Cs$ and the $D$. The DA and CA in the PTGS are performed in an alternative way during training.

Notations used in this paper are explained as follows. We denote the segmentation model $h$ as $h^{\theta, \phi}(x) = C^{\theta}(G^{\phi}(x))$ which consists of a feature extractor $G^{\phi}$ parameterized by $\phi$ and a classifier $C^{\theta}$ parameterized by $\theta$, and $x$ is a sample from $\mathcal{U}_s$ or $\mathcal{U}_t$. If multiple classifiers are used, we denote the $j^{th}$ classifier as $C_{j}$. We denote the discriminator as $D^{\psi}$ parameterized by $\psi$.

\subsection{Bounds Relation}
\label{sec:bounds_relation}
We start by examining the relationship between the DA and the CA from the perspective of target error bound. We propose to use this relation to improve the segmentation performance of class alignment, which is desired for dense prediction tasks. 
We provide the following theorem: 

\textbf{Theorem 2} If we assume there is a hypothesis space $\mathcal{H}$ for segmentation model $h^{\theta, \phi}$ and a hypothesis space $\mathcal{H}_D$ for domain classifiers $D^\psi$, and $\mathcal{H}\Delta\mathcal{H}\subset \mathcal{H}_D$, then we have
\begin{equation}
    \label{eq:bound_relation}
    \begin{aligned}
        \epsilon_t(h) &\leq \hat{\mathbb{UB}}_2 \leq \hat{\mathbb{UB}}_1,\\
        \hat{\mathbb{UB}}_1 &= \epsilon_s(h)+\frac{1}{2}d_{\mathcal{H}_D}(\mathcal{D}_s, \mathcal{D}_t)+\lambda,\\
        \hat{\mathbb{UB}}_2 &= \epsilon_s(h)+\frac{1}{2}d_{\mathcal{H}\Delta\mathcal{H}}(\mathcal{D}_s, \mathcal{D}_t)+\lambda.
    \end{aligned}
\end{equation}
The proof of this theorem is provided in Section.~\ref{sec:proof}.

Essentially, we limit the hypothesis space $\mathcal{H}$ and $\mathcal{H}_D$ in Eq.~(\ref{eq:bound_relation}) into the space of deep neural networks. Directly optimizing over $\hat{\mathbb{UB}_2}$ might be hard to converge since $\hat{\mathbb{UB}_2}$ is a tighter upper bound for the prediction error on the target domain.  
The bounds relation in Eq.~(\ref{eq:bound_relation}) shows that the $\hat{\mathbb{UB}}_1$ is an upper bound of $\hat{\mathbb{UB}}_2$. This provides us a clue to improve the training process of class alignment, i.e.,  \textit{the domain alignment can be a global constraint and narrow down the searching space for the class alignment}. This also implies that integrating the domain alignment and class alignment might boost the training efficiency as well as the prediction performance of UDA. This inspires us to design a new model, which we describe in the subsequent sections. 

\subsection{Model Structure }
\label{sec:structure}
Following our proposed PTGS and the identified relation in Eq. (\ref{eq:bound_relation}), we design the structure of our CALI model as shown in stage I of Fig.~\ref{fig:net}. Four networks are involved, a shared feature extractor $G^{\phi}$, a domain discriminator $D^{\psi}$ and two classifiers $C_1^{\theta}$ and $C_2^{\theta}$. 
Furthermore, as defined in Eq.~(\ref{eq:hdh_distance}), $h$ and $h^{'}$ are two different hypotheses, thus we have to ensure the classifiers $C_1^{\theta}$ and $C_2^{\theta}$ are different. Note that the supervision signal from the source domain ($\mathbf{y}_s$ in Fig.~\ref{fig:net}) is used to train $C_1^{\theta}$ and $C_2^{\theta}$ because both classifiers are expected to generate correct decision boundaries on the source domain. No label from the target domain is used during training.

We also show steps involved in our ICALI model in Fig. \ref{fig:net}-stages II \& III. Stage II shows how a pair of a mixed image and the corresponding mixed label is generated. First, the prediction of the source image $\mathbf{o}_{1s}$ is used to compute the performance (indicated by the mean Intersection over Union (mIoU)) for all classes. Note that $\mathbf{o}_{1s}$ is the prediction of $C_1^{\theta}$ in stage I. The classes are then divided into a group of well-performing classes and a group of under-performing classes. The ratio for the division is a hyperparameter. Then a selection mask for source data $\mathbf{M}_s$ is generated by extracting the regions of under-performing classes in the source ground-truth label. We can easily obtain the mask for target data: $\mathbf{M}_t = \mathbf{1} - \mathbf{M}_s$. Next, the mixed data can be generated by:
\begin{equation}
    \label{eq:mixed_data}
    \begin{aligned}
        &\mathbf{x}_m = (\mathbf{x}_s\odot \mathbf{M}_s) \oplus (\mathbf{x}_t\odot \mathbf{M}_t), \\
        & \mathbf{y}_m = (\mathbf{y}_s\odot \mathbf{M}_s) \oplus (\mathbf{o}_{1t}\odot \mathbf{M}_t),
    \end{aligned}
\end{equation}
where $\mathbf{o}_{1t}$ is the prediction of $C_1^{\theta}$ in stage I.

In stage III, the newly generated data $ \left\{ \mathbf{x}_m, \mathbf{y}_m \right\}$ are used to train the model $G^{\psi}$ and $C_1^\theta$. Note that the model $C_2^\theta$ is not included in the training of stage III such that the difference between the two models can be well maintained. 

\subsection{Losses}
We denote raw images from the source or target domain as $\mathbf{x}$, and the label from the source domain as $\mathbf{y}$. We use semantic labels in the source domain to train all of the nets, but the domain discriminator, in a supervised way (see the solid red one-way arrow in Fig.~\ref{fig:net}). We need to minimize the supervised segmentation loss since Eq.~(\ref{eq:bound_relation}) and other related Eqs suggest that the source prediction error is also part of the upper bound of the target error. In this section, we omit superscripts for all model notations. The supervised segmentation loss for training CALI is defined as 
\begin{equation}
    \label{eq:seg_loss}
    \begin{aligned}
    &\mathcal{L}_{seg} (G, C_1, C_2) = \frac{1}{2}\Bigg( \Bigg. \mathbb{E}_{(\mathbf{x}, \mathbf{y})\sim \mathcal{D}_s} \left[ -\mathbf{y}\log (C_1(G(\mathbf{x}))) \right] + \\
    &~~~~~~~~~~~~~~~~~~~~~~~~~~~\mathbb{E}_ {(\mathbf{x}, \mathbf{y})\sim \mathcal{D}_s} \left[ -\mathbf{y}\log (C_2(G(\mathbf{x}))) \right]\Bigg. \Bigg)\\
    &= -\frac{1}{2}\mathbb{E}_{(\mathbf{x}, \mathbf{y})\sim \mathcal{D}_s} [\mathbf{y}_s \log \left( (C_1(G(\mathbf{x}))\odot (C_2(G(\mathbf{x})))) \right)],
    \end{aligned}
\end{equation}
where $\odot$ represents the element-wise multiplication between two tensors.

To perform domain alignment, we need to define the joint loss function for $G$ and $D$ 
\begin{equation}
    \label{eq:domain_loss}
        \mathcal{V}_1(G, D) = -\left(  \mathcal{CE}_s(\mathbf{x}) + \mathcal{CE}_t(\mathbf{x})\right),
\end{equation}
where no segmentation labels but domain labels are used, and we use the standard cross-entropy to compute the domain classification loss for both source ($\mathcal{CE}_s(\mathbf{x})$) and target data ($\mathcal{CE}_t(\mathbf{x})$). We have
\begin{equation}
    \label{eq:source_ce}
    \begin{aligned}
    \mathcal{CE}_s(\mathbf{x}) &= \mathbb{E}_{\mathbf{x}\sim \mathcal{D}_s} [\mathcal{CE}([1, 0]^T, [D(G(\mathbf{x})), 1-D(G(\mathbf{x}))]^T)]\\
    &= \mathbb{E}_{\mathbf{x}\sim \mathcal{D}_s} [-\log (D(G(\mathbf{x})))].
    \end{aligned}
\end{equation}
and
\begin{equation}
    \label{eq:target_ce}
    \begin{aligned}
    \mathcal{CE}_t(\mathbf{x}) &= \mathbb{E}_{\mathbf{x}\sim \mathcal{D}_t} [\mathcal{CE}([0, 1]^T, [D(G(\mathbf{x})), 1-D(G(\mathbf{x}))]^T)]\\
    &= \mathbb{E}_{\mathbf{x}\sim \mathcal{D}_t}[ -\log (1-D(G(\mathbf{x})))],
    \end{aligned}
\end{equation}
Note we include $G$ in Eq.~(\ref{eq:source_ce}) since both the source data and target data are passed through the feature extractor. This is different than standard GAN, where the real data is directly fed to $D$, without passing through the generator.

\begin{algorithm}[t]
\label{algo:training}
\caption{\small Training Process}
\begin{algorithmic}[1]
{\small
\STATE \textbf{Input:} Source dataset $\mathcal{U}_s$; Target dataset $\mathcal{U}_t$; Initial model $G, C_1, C_2$ and $D$; Maximum iterations $M$; Iteration interval $I$.
\STATE \textbf{Output:} Updated model parameters $\phi_{G}, \theta_{C_1}, \theta_{C_2}$ and $\psi_{D}$.
\vspace*{-9pt}
\STATE \textbf{Initialization:} \textit{is\_domain=True; is\_class=False;}
\FOR{m $\leftarrow$ 1~to~M}
    {   
        \IF{$m \% I==0$ and $m \neq 0$}
          {
           \STATE is\_domain = not is\_domain;\\
           \STATE is\_class = not is\_class;\\
          }
        \ENDIF
        \STATE {\em /* Eq.~(\ref{eq:seg_loss}) */}
        \STATE { $\min_{\phi_{G}, \theta_{C_1}, \theta_{C_2}} \mathcal{L}_{seg}(G, C_1, C_2);$ }
        \STATE {\em /* Eq.~(\ref{eq:structural_loss}) */}
        \STATE {$\min_{\theta_{C_1}, \theta_{C_2}} \mathcal{WR}(C_1, C_2);$}
        \IF{is\_domain}{
            \STATE {\em /* Eq.~(\ref{eq:domain_loss}) */}
            \STATE {$\max_{\psi_{D}} \min_{\theta_{G}} \mathcal{V}_1(G, D);$}
        }
        \ENDIF
        \IF{is\_class}{
            \STATE {\em /* Eq.~(\ref{eq:class_loss}) */}
            \STATE {$\max_{\theta_{C_1}, \theta_{C_2}} \min_{\phi_{G}} \mathcal{V}_2(G, C_1, C_2);$}
        }
        \ENDIF
        \STATE {\em /* Only for ICALI - Eq. (\ref{eq:mixed_loss}) */}
        \STATE { $\min_{\phi_{G}, \theta_{C_1}} \mathcal{L}_{m}(G, C_1);$ }
    }
\ENDFOR
\STATE Return $\phi_{G}$, $\theta_{C_1}$, $\theta_{C_2}$ and $\psi_{D}$;
} 
\end{algorithmic}
\end{algorithm}

To perform class alignment, we need to define the joint loss function for $G$, $C_1$, and $C_2$
\begin{equation}
    \label{eq:class_loss}
    \begin{aligned}
    &\mathcal{V}_2(G, C_1, C_2) = \mathbb{E}_{\mathbf{x}\sim \mathcal{D}_t}\left [ d(C_1(G(\mathbf{x})), C_2(G(\mathbf{x}))) \right ],
    \end{aligned}
\end{equation}
where $d(\cdot, \cdot)$ is the distance measure between two distributions from the two classifiers. In this paper, we use the same $L_1$ distance in \citet{saito2018maximum} as the measure, thus $d(p, q) = \frac{1}{K}|p-q|_1$, where $p$ and $q$ are two distributions and $K$ is the number of label classes. 

To prevent $C_1$ and $C_2$ from converging to the same network throughout the training, we use the cosine similarity as a weight regularization to maximize the difference of the weights from $C_1$ and $C_2$, i.e.,  
\begin{equation}
    \label{eq:structural_loss}
    \mathcal{WR}(C_1, C_2)=\frac{\mathbf{w}_1\cdot \mathbf{w}_2}{\left\|\mathbf{w}_1 \right\|\left\|\mathbf{w}_2 \right\|},
\end{equation}
where $\mathbf{w}_1$ and $\mathbf{w}_2$ are the weight vectors of $C_1$ and $C_2$, respectively. 

The extra supervised training in ICALI only involves a simple loss:
\begin{equation}
    \label{eq:mixed_loss}
    \mathcal{L}_m = \mathbb{E}_{(\mathbf{x}, \mathbf{y})\sim \mathcal{D}_m}[-\mathbf{y}\log (C_1(G({\mathbf{x})}))].
\end{equation}

\subsection{Training Algorithm}
We integrate the training processes of domain alignment and class alignment to systematically train our CALI model. To be consistent with  Eq.~(\ref{eq:bound_relation}), we adopt an iterative mechanism that alternates between domain alignment and class alignment.
We present the pseudo-code for the training process of CALI and ICALI in Algorithm 4.1.

Note the adversarial training order of $\mathcal{V}_1$ in Algorithm 4.1 is $\max_{\psi_D} \min_{\phi_G}$, instead of the $\min_{\phi_G} \max_{\psi_D}$, meaning in each training iteration we first train the feature extractor and then the discriminator. The reason for this order is because we empirically find that the feature from $G$ is relatively easy for $D$ to discriminate, hence if we train $D$ first, then the $D$ might become an accurate discriminator in the early stage of training and there will be no adversarial signals for training $G$, thus making the whole training fail. The same order applies to training of the pair of $G$ and $Cs$ with $\mathcal{V}_2$.

\begin{figure}
{
  \centering
    \subfigure[]
  	{\label{fig:cv_real}\includegraphics[width=0.3\linewidth]{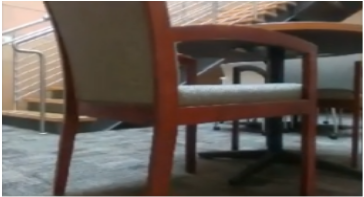}}
    \subfigure[]
  	{\label{fig:cv_sn}\includegraphics[width=0.3\linewidth]{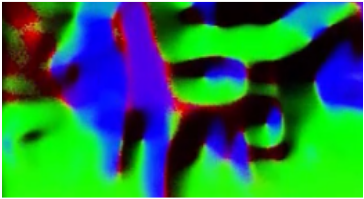}}
   	\subfigure[]
   	{\label{fig:cv_seg}\includegraphics[width=0.3\linewidth]{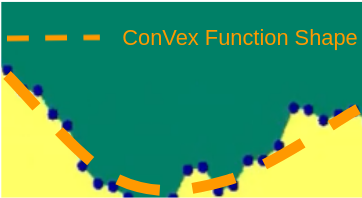}}
   	\subfigure[]
   	{\label{fig:cc_real}\includegraphics[width=0.3\linewidth]{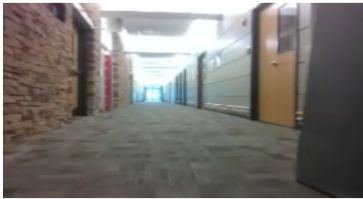}}
   	\subfigure[]
   	{\label{fig:cc_sn}\includegraphics[width=0.3\linewidth]{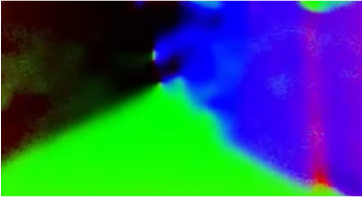}}
   	\subfigure[]
   	{\label{fig:cc_seg}\includegraphics[width=0.3\linewidth]{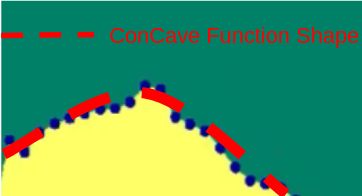}} \vspace{-10pt}
   \caption{\small Images from the camera on a ground robot in a real indoor environment. \textit{Left column}: RGB images; \textit{Middle column}: Surface normal images; \textit{Right column}: Predicted segmentation images (yellow: navigable; green: non-navigable). Blue dots are sampled points from the predicted boundary. The orange dash line represents an approximated convex function shape while the red dash line is an approximated concave function shape. \vspace{-15pt}
   }
  }
\label{fig:convex}  
\end{figure}

\subsection{Visual Planner}
We design a visual receding horizon planner to achieve feasible visual navigation by combining the learned image segmentation. 
Specifically, first we compute a library of motion primitives~\citep{howard2007optimal, howard2008state} $\mathcal{M} = \left \{ \mathbf{p}_1, \mathbf{p}_2, \cdots, \mathbf{p}_n \right \}$ 
where each $\mathbf{p}_* = \left \{ \mathbf{x}_1, \mathbf{x}_2, \cdots, \mathbf{x}_m \right \}$ is a single primitive. 
We use $\mathbf{x}_{*} = \begin{bmatrix}
x & y & \psi
\end{bmatrix}^T$ to denote a robot pose. 
Then we project the motion primitives to the image plane and compute the navigation cost function for each primitive based on the evaluation of collision risk in image space and target progress. Finally, we select the primitive with minimal cost 
to execute. The trajectory selection problem can be defined as:
\begin{equation}
    \label{eq:traj_selection}
    \mathbf{p}_{optimal} = \underset{\mathbf{p}}{\text{argmin}}~ w_1 \cdot C_{c}(\mathbf{p}) + w_2 \cdot C_{t}(\mathbf{p}),
\end{equation}
where $C_{c}(\mathbf{p}) = \sum_j^m c_c^j$ and $C_{t}(\mathbf{p}) = \sum_j^m c_t^j$ are the collision cost and target cost of one primitive $\mathbf{p}$, and $w_1$, $w_2$ are corresponding weights, respectively.

\begin{figure}
  \centering
    \subfigure[]
  	{\label{fig:surface_normal}\includegraphics[width=0.3\linewidth]{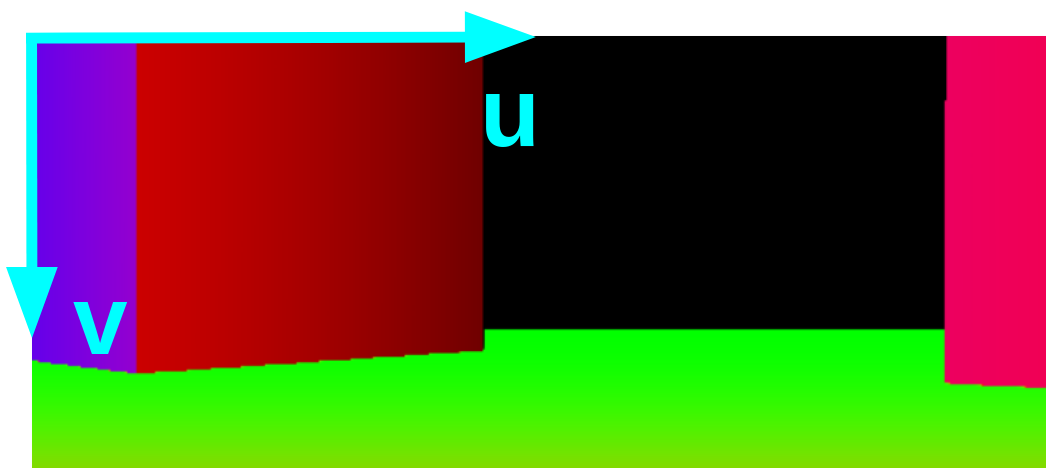}}
    \subfigure[]
  	{\label{fig:segmentation}\includegraphics[width=0.3\linewidth]{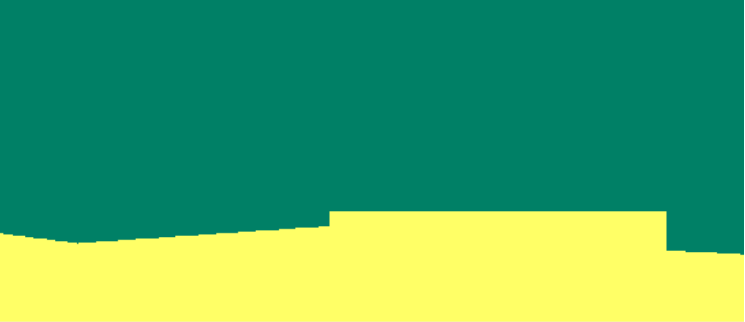}}
   	\subfigure[]
   	{\label{fig:10}\includegraphics[width=0.3\linewidth]{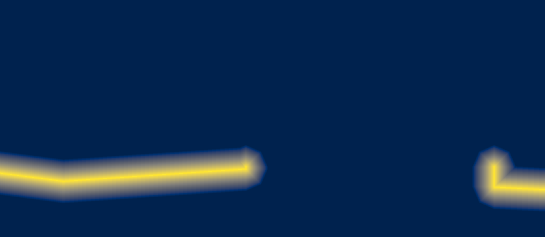}}
   	\subfigure[]
   	{\label{fig:25}\includegraphics[width=0.3\linewidth]{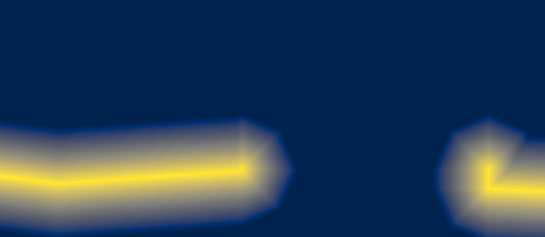}}
   	\subfigure[]
   	{\label{fig:55}\includegraphics[width=0.3\linewidth]{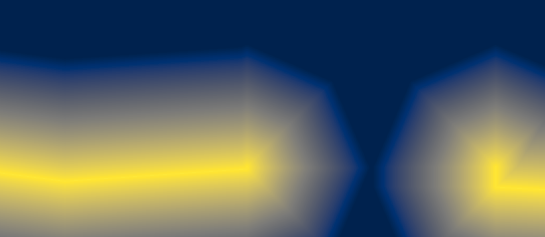}}
   	\subfigure[]
   	{\label{fig:100}\includegraphics[width=0.3\linewidth]{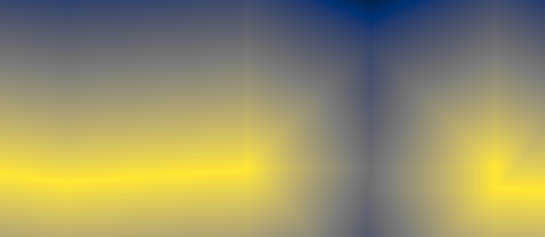}} \vspace{-10pt}
   \caption{\small Illustrations in Gazebo simulation. (a)~Surface normal image; (b)~Binary navigable space segmentation and different EDFs with varying scale factors of (c)~$\alpha=0.05$, (d)~$\alpha=0.25$, (e)~$\alpha=0.55$ and (f)~$\alpha=1.00$. \vspace{-10pt}
   }\label{fig:sedf}  
 \end{figure}

\subsubsection{Collision Avoidance}
In this work, we propose a Scaled Euclidean Distance Field~(SEDF) for obstacle avoidance. Conventional collision avoidance is usually conducted in the map space\\\citep{gao2017gradient, han2019fiesta}, where an occupancy map and the corresponding Euclidean Signed Distance Field~(ESDF) have to be provided in advance or constructed incrementally in real time. In this work instead we eliminate this expensive map construction process and evaluate the collision risk directly in the image space. Specifically, we first compute a SEDF image $E^{'}(S)$ based on an edge map $Edge(S)$ detected in the learned binary segmentation $S(I)$, where $I$ is the input image. We then project the motion primitives from the map space to the image space and evaluate all primitives' projections in $E^{'}(S)$. 

To perform obstacle avoidance in image space, we have to detect the obstacle boundary in $Edge(S)$. To achieve this, we propose to categorize the edges $Edge(S)$ into two classes, \textit{Strong Obstacle Boundaries~(SOBs)} and \textit{Weak Obstacle Boundaries~(WOBs)}. We treat the boundary from the binary segmentation as a function of a single variable in image space, and we use the twin notions of convexity and concavity of functions to define the SOBs and WOBs, respectively. SOBs mean obstacles are near to the robot (e.g., the random furniture closely surrounding the robot) and they cause the boundaries to exhibit an approximated \textit{ConVex Function Shape~(CVFS)} (see Fig.~\ref{fig:cv_seg}). WOBs indicate obstacles are far from the robot 
(e.g., the wall boundaries that the robot makes large clearance from) 
and they typically make the boundaries reveal an approximated \textit{ConCave Function Shape~(CCFS)} (see Fig.~\ref{fig:cc_seg}). In this work, we only consider the obstacles with SOBs and adopt a straightforward way (as in Eq.~(\ref{eq:boundary_set})) to detect boundary segments in $Edge(S)$ with CVFS. We use a points set $\Omega$ to represent the boundary segments:
\begin{equation}
    \label{eq:boundary_set}
    \Omega = \left \{ (u, v), (u, v)\in Edge(S)~\text{and}~v > v_{thres} \right \},
\end{equation}
where $(u, v)$ are the coordinates in the image frame, as shown in Fig.~\ref{fig:surface_normal},  and $v_{thres}$ is a pre-defined value for evaluating the boundary convexity. If we use $\partial \Omega$ to denote the boundary of a set, then in our case we have $\partial \Omega = \Omega$. Then the definition of an EDF is:
\begin{equation}
    \label{eq:edf}
    \begin{aligned}
        E[ u, v ] &= d((u, v), \partial \Omega)\\
        d(x, \partial \Omega) &\coloneqq \underset{y\in \partial \Omega}{\inf} d(x, y),
    \end{aligned}
\end{equation}
where $d(x, y) = \left \| x - y \right \|$ is the Euclidean distance between vectors $x$ and $y$.

However, directly computing an EDF using Eq.~(\ref{eq:edf}) in the image space will propagate the obstacles' gradients to the whole image space, which might cause the planning evaluation space to be too limited. To address this, we introduce a scale factor $\alpha$ to compute a corrected version of EDF:
\begin{equation}
    \label{eq:corrected_edf}
        E^{'}[u, v]= 
        \begin{dcases}
            E[u, v],& \text{if } E[u, v] \leq \alpha\cdot d_{max}\\
            \alpha\cdot d_{max},              & \text{otherwise}
        \end{dcases}
\end{equation}
where $\alpha\in [0, 1]$, and $d_{max} = \underset{u\in \mathcal{U}, v\in \mathcal{V}}{max}E[u, v]$, where $\mathcal{U}$ and $\mathcal{V}$ are the rows and columns index sets, respectively. Some examples of $E^{'}$ with different $\alpha$ values can be seen in Fig.~\ref{fig:sedf}.

Assuming $\mathbf{x}^j$ is the $j^{th}$ pose in one primitive and its image coordinates are $\left ( u^j, v^j \right )$, then the collision risk for $\mathbf{x}^j$ is
\begin{equation}
    \label{eq:collision_cost}
        c_c^j = E^{'}[u^j, v^j].
\end{equation}

\subsubsection{Target Progress}
To evaluate target progress during the navigation pro-gress, we propose to use the distance on $SE(3)$ as the metric. We define three types of frames: world frame $F_w$, primitive pose frame $F_{pj}$, and goal frame $F_g$. The transformation of $F_{pj}$ in $F_w$ is denoted as $\mathbf{T}_{wpj}$ while that of $F_g$ in $F_w$ is $\mathbf{T}_{wg}$. A typical approach to represent the distance is to split a pose into a position and an orientation and define two distances on $\mathbb{R}^3$ and $SO(3)$. Then the two distances can be fused in a weighted manner with two strictly positive scaling factors $a$ and $b$ and with an exponent parameter $p\in [1, \infty]$~\citep{bregier2018defining}:
\begin{equation}
    \label{eq:se3_dist_1}
    \begin{aligned}
        d(\mathbf{T}_{wpj}, \mathbf{T}_{wg}) = \Bigg[ \Bigg. &a \cdot d_{rot}(\mathbf{R}_{wpj}, \mathbf{R}_{wg})^p+\\
        &b \cdot d_{trans}(\mathbf{t}_{wpj}, \mathbf{t}_{wg})^p\Bigg. \Bigg]^{1/p}.
    \end{aligned}
\end{equation}
We use the Euclidean distance as $d_{trans}(\mathbf{t}_{wpj}, \mathbf{t}_{wg})$, the Riemannian distance over $SO(3)$ as $d_{rot}(\mathbf{R}_{wpj}, \mathbf{R}_{wg})$ and set $p$ as $2$. Then the distance (target cost) between two transformation matrices can be defined~\citep{park1995distance} as:
\begin{equation}
\label{eq:se3_dist_2}
\begin{aligned}
    c_t^j &= d(\mathbf{T}_{wpj}, \mathbf{T}_{wg}) \\
          &= \left [ a \cdot \left \| \log(\mathbf{R}_{wpj}^{-1}\mathbf{R}_{wg}) \right \|^2+b \cdot \left \| \mathbf{t}_{wpj} - \mathbf{t}_{wg} \right \|^2 \right ]^{1/2}.
\end{aligned}
\end{equation}

\section{Experiments}
\label{sec:experiments}
\subsection{Datasets}
\begin{figure*}
{
\centering
  {\includegraphics[width=1\linewidth]{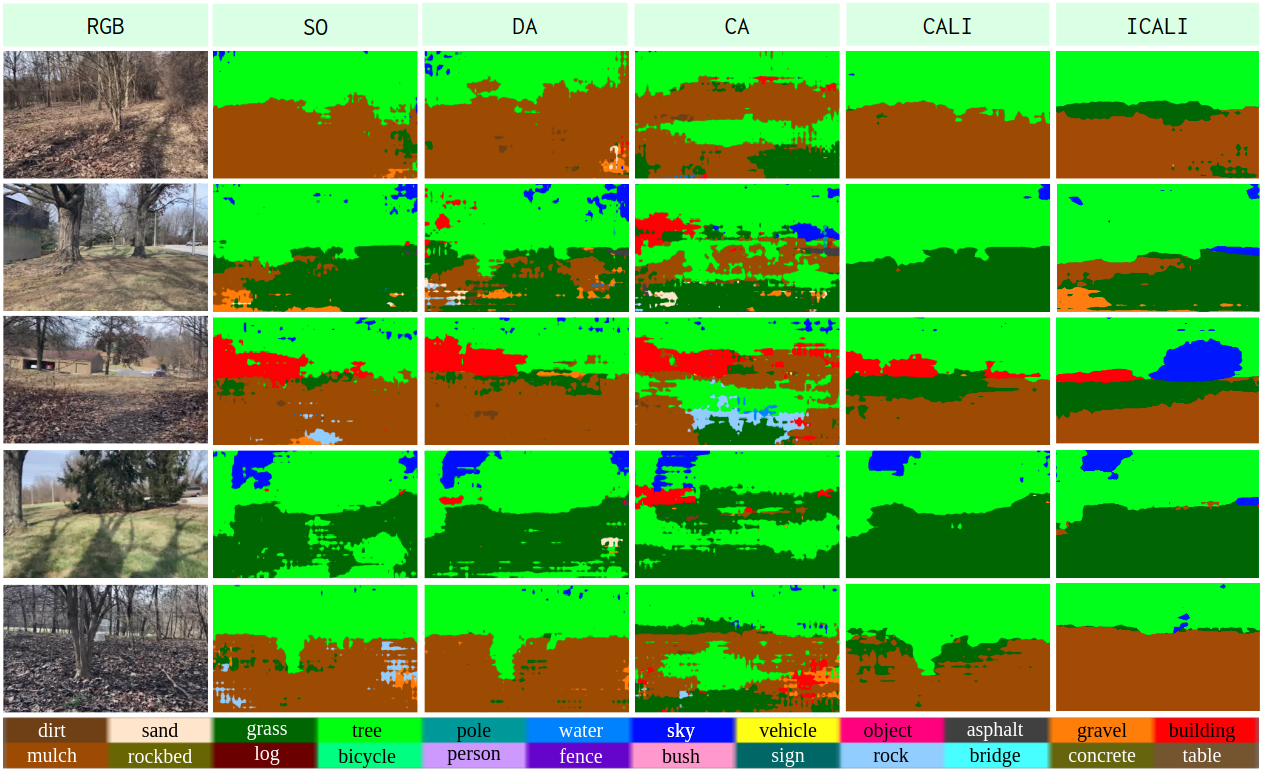} 
  } 
\caption{\small Qualitative results on adaptation RUGD$\rightarrow$MESH. Results of our proposed model are listed in the last (ICALI) and last second (CALI) columns.
}  
\label{fig:mesh_images} 
}
\end{figure*}

We evaluate CALI together with several baseline methods on a few challenging domain adaptation scenarios, where several public datasets, e.g., GTA5~\citep{richter2016playing}, Cityscapes~\citep{Cordts2016Cityscapes}, RUGD~\citep{RUGD2019IROS}, RELLIS~\citep{jiang2020rellis3d}, as well as a small self-collected dataset, named MESH (see the first column of Fig.~\ref{fig:mesh_images}), are investigated. 
The GTA5 dataset contains 24966 synthesized high-resolution images in the urban environments from a video game and pixel-wise semantic annotations of 33 classes. 
The Cityscapes dataset consists of 5000 finely annotated images whose label is given for 19 commonly seen categories in urban environments, e.g., road, sidewalk, tree, person, car, etc. 
The RUGD and RELLIS are two datasets that aim to evaluate segmentation performance in off-road environments. 
The RUGD and the RELLIS contain 24 and 20 classes with 8000 and 6000 images, respectively. RUGD and RELLIS cover various scenes like trails, creeks, parks, villages, and puddle terrains. 
Our dataset, MESH, includes features like grass, trees (particularly challenging in winter due to foliage loss and monochromatic colors), mulch, etc. It helps us to further validate the performance of our proposed model for traversability prediction in challenging scenes, particularly the off-road environments.

\subsection{Implementation Details}
To be consistent with our theoretical analysis, the implementation of CALI only adopts the necessary indications by Eq.~(\ref{eq:bound_relation}). First, Eq.~(\ref{eq:bound_relation}) requires that the input of the two upper bounds (one for DA and the other one for CA) should be the same. Second, nothing else but only domain classification and hypotheses discrepancy are involved in Eq.~(\ref{eq:bound_relation}) and other related analyses (Eq.~(\ref{eq:theorem1}) - Eq.~(\ref{eq:ub2})). Accordingly, we strictly follow the guidance of our theoretical analyses. First, CALI performs DA in the  intermediate-feature level ($f$ in Fig.~\ref{fig:net}), instead of the output-feature level used in \citet{vu2019advent}. Second, we exclude the multiple additional tricks, e.g., entropy-based and multi-level features based alignment, and class-ratio priors in \citet{vu2019advent} and multi-steps training for feature extractor in \citet{saito2018maximum}. We also implement baseline methods without those techniques for a fair comparison. To avoid possible degraded performance brought by a class imbalance in the used datasets, we regroup those rare classes into classes with a higher pixel ratio. For example, we treat the building, wall, and fence as the same class; the person and rider as the same class in the adaptation of GTA5$\rightarrow$Cityscapes. In the adaptation of RUGD$\rightarrow$RELLIS, we treat the tree, bush, and log as the same class, and the rock and rockbed as the same class. Details about remapping can be seen in Fig.~\ref{fig:city_cloud} and Fig.~\ref{fig:outdoor_cloud} in Section.~\ref{sec:remapping}.

\begin{figure*}
{
\centering
  {\includegraphics[width=\linewidth]{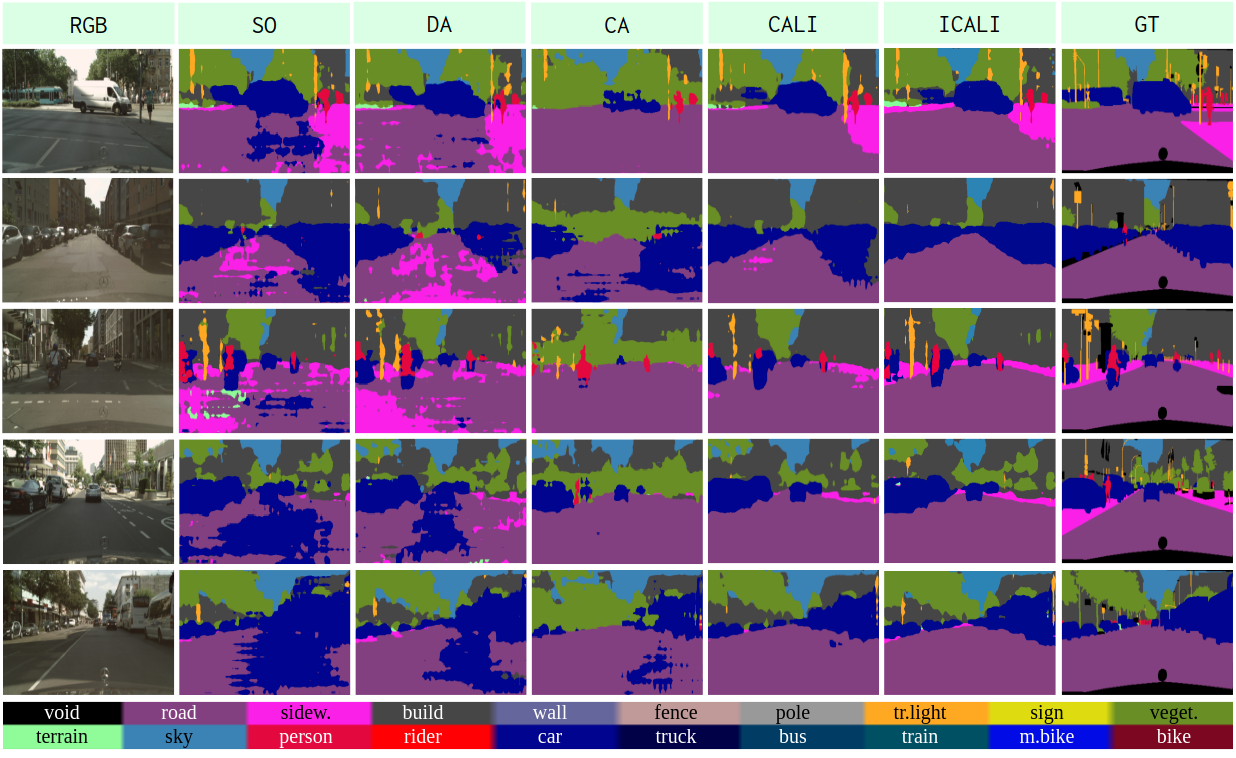} 
  } 
\caption{\small Qualitative results on adaptation GTA5$\rightarrow$Cityscapes. Results of our proposed model are listed in the last second (ICALI) and third (CALI) columns. GT represents the ground-truth labels.}
\label{fig:city_images} 
}
\end{figure*}

\begin{figure*}
{
\centering
  {\includegraphics[width=\linewidth]{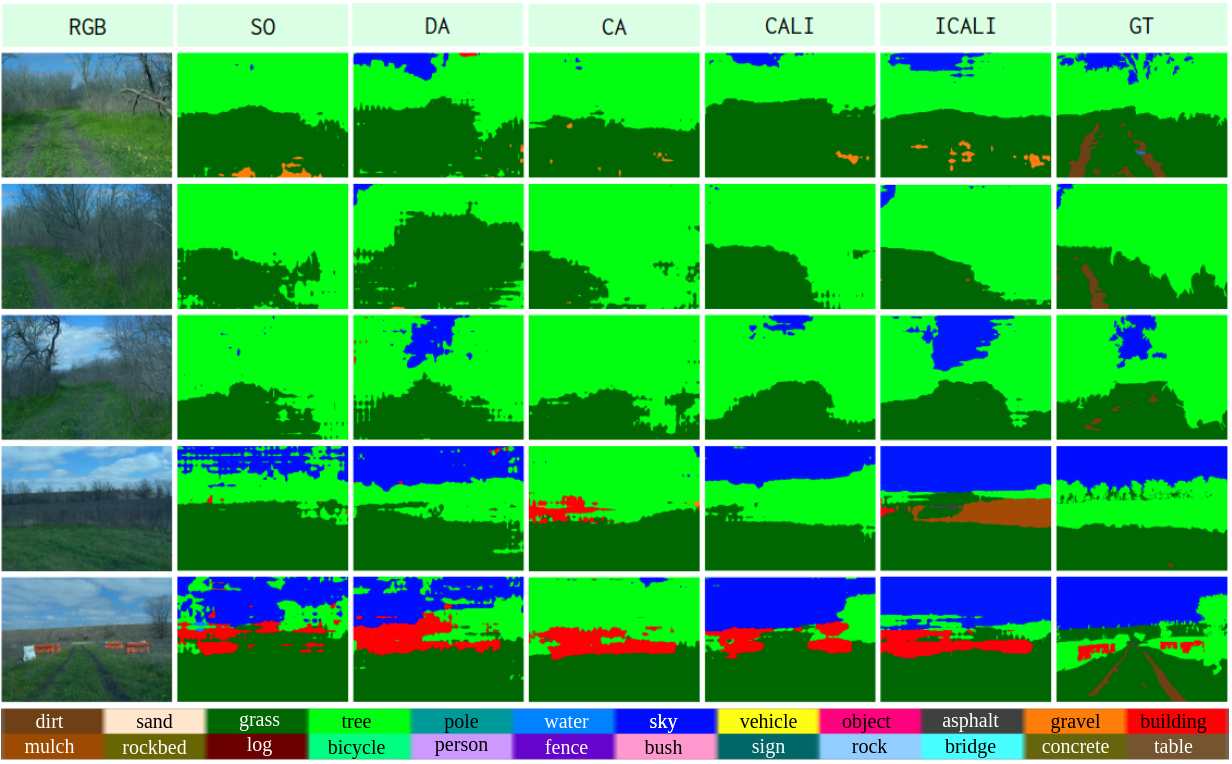} 
  } 
\caption{\small Qualitative results on adaptation RUGD$\rightarrow$RELLIS. Results of our proposed model are listed in the last second (ICALI) and third (CALI) columns. GT represents the ground-truth labels.}
\label{fig:outdoor_images} 
}
\end{figure*}

We use the PyTorch~\citep{paszke2019pytorch} framework for implementation.  Training images from source and target domains are cropped to be half of their original image dimensions. The batch size is set to 1 and the weights of all batch normalization layers are fixed. We use the ResNet-101~\citep{he2016deep} pretrained on ImageNet~\citep{deng2009imagenet} as the model $G$ for extracting features. We use the ASPP module in DeepLab-V2~\citep{chen2017deeplab} as the structure for $C_1$ and $C_2$. We use the similar structure in \citet{radford2015unsupervised} as the discriminator $D$, which consists of 5 convolution layers with kernel $4\times 4$ and with channel size $\left\{ 64, 128, 256, 512, 1 \right\}$ and stride of 2. Each convolution layer is followed by a Leaky-ReLU~\citep{maas2013rectifier} parameterized by 0.2, but only the last convolution layer is follwed by a Sigmoid function. During the training, we use SGD~\citep{bottou2010large} as the optimizer for $G, C_1$ and $C_2$ with a momentum of 0.9, and use Adam~\citep{kingma2014adam} to optimize $D$ with $\beta_1=0.9, \beta_2=0.99$. We set all SGD optimizers a weight decay of $5\text{e-}4$. The initial learning rates of all SGDs for performing domain alignment are set to $2.5\text{e-}4$ and the one of Adam is set as $1\text{e-}4$. For class alignment, the initial learning rate of SGDs is set to $1\text{e-}3$. All of the learning rates are decayed by a poly learning rate policy, where the initial learning rate is multiplied by $(1-\frac{iter}{max\_iters})^{power}$ with $power=0.9$. All experiments are conducted on a single Nvidia Geforce RTX 2080 Super GPU.

\subsection{Comparative Studies}

We present comparative experimental results of our proposed model, CALI, compared to different baseline methods -- Source-Only (SO) method, Domain-Alignment (DA)~\citep{vu2019advent} method, and Class-Alignment\\\citep{saito2018maximum} method. Specifically, we first perform evaluations on a sim2real UDA in city-like environments, where the source domain is represented by GTA5 while the target domain is the Cityscapes. Then we consider a transfer of real2real in forest environments, where the source domain and target domain are set as RUGD and RELLIS, respectively. All models are trained with full access to the images and labels in the source domain and with only access to the images in the target domain. The labels in target datasets are only used for evaluation purposes. Finally, we further validate our model performance for adapting from RUGD to our self-collected dataset MESH. 

To ensure a fair comparison, all the methods use the same feature extractor $G$; both DA and CALI have the same domain discriminator $D$; both CA and CALI have the same two classifiers $C_1$ and $C_2$. We also use the same optimizers and optimization-related hyperparameters if any is used for models under comparison. 

We use the mIoU as the metric to evaluate each class and overall segmentation performance on testing images. IoU is computed as $\frac{n_{tp}}{n_{tp} + n_{fp} + n_{fn}}$, where $n_{tp}, n_{tn}, n_{fp}$ and $n_{fn}$ are true positive, true negative, false positive and false negative, respectively.

\subsubsection{GTA5$\rightarrow$Cityscapes}
Quantitative comparison results of GTA5$\rightarrow$Cityscapes are shown in Table~\ref{tb:city_quantitative}, where segmentations are evaluated on 9 classes (as regrouped in Fig.~\ref{fig:city_cloud}). Our proposed methods (CALI \& ICALI) have significant advantages over multiple baseline methods, and ICALI achieves the best performance for all categories and overall performance (mIoU*).

In our testing case, SO achieves the highest score for the class person even without any domain adaptation. One possible reason for this is the deep features of the source person and the target person from the model solely trained on source domain, are already well-aligned. If we try to interfere this well-aligned relation using \textit{unnecessary} additional efforts, the target prediction error might be increased (see the mIoU values of the person from the other three methods). We call this phenomenon as \textit{negative transfer}, which also happens to other classes if we compare SO and DA/CA, e.g., sidewalk, building, sky, vegetation, and so on. In contrast, CALI maintains an improved performance compared to either SO or DA/CA. We validate our analytical method for DA and CA (Section~\ref{sec:bounds_relation}) by a comparison between CALI and baselines. This indicates either single DA or CA is problematic for semantic segmentation, particularly when we strictly follow what the theory supports and do not include any other training tricks (that might increase the training complexity and make the training unstable). This implies that integration of DA and CA is beneficial to each other with significant improvements, and more importantly, CALI is well theoretically supported, and the training process is easy and stable. Based on CALI, ICALI further improves performance and achieves the best results for all classes, validating the effectiveness of introducing the extra training of mixed data into CALI.

Fig.~\ref{fig:city_images} shows the examples of qualitative comparison for UDA of GTA5$\rightarrow$Cityscapes. We find that CALI prediction is less noisy compared to the baseline methods as shown in the second and third columns (sidewalk or car on-road), and shows better completeness (part of the car is missing, see the fourth column). ICALI further improves the segmentation quality, especially for challenging classes, e.g., sidewalk, traffic light, and person.

{
\begin{table}
\caption{\small Quantitative comparison of different methods in UDA of GTA5$\rightarrow$Cityscapes. mIoU* represents the average mIoU over all of classes.}
\centering 
{
\begin{tabularx}{0.96\linewidth}{cccc|cc}
\hline\hline 
  Class & SO & DA & CA & CALI & ICALI \\[0.5ex]
\hline 
Road & 38.86 & 52.80 & 78.56 & 75.36 & \textbf{84.7} \\ [1ex]
Sidewalk & 17.47 & 18.95 & 2.79 & 27.12 & \textbf{39.91} \\ [1ex]
Building & 63.60 & 61.73 & 43.51 & 67.00 & \textbf{71.14} \\ [1ex]
Sky & 58.08 & 54.35 & 46.59 & 60.49 & \textbf{64.08} \\ [1ex]
Vegetation & 67.21 & 64.69 & 41.48 & 67.50 & \textbf{72.14} \\ [1ex]
Terrain & 7.63 & 7.04 & 8.37 & 9.56 & \textbf{11.12}\\ [1ex]
Person & 16.89 & 15.45 & 13.48 & 15.03 & \textbf{17.71}\\ [1ex]
Car & 30.32 & 43.41 & 31.64 & 52.25 & \textbf{62.49}\\ [1ex]
Pole & 11.61 & 12.38 & 9.68 & 11.91 & \textbf{15.48}\\ [1ex]
mIoU* & 34.63 & 36.76 & 30.68 & 42.91 & \textbf{48.75} \\
\hline\hline 
\end{tabularx}\vspace{-10pt}
}
\label{tb:city_quantitative} 
\end{table}
}

\subsubsection{RUGD$\rightarrow$RELLIS}
We show quantitative results of RUGD$\rightarrow$RELLIS in Table~\ref{tb:outdoor_quantitative}, where only 5 classes $^{\ddag}$ are evaluated.
\footnotetext{$^{\ddag}$ This is because other classes (in Fig.~\ref{fig:outdoor_cloud}) frequently appearing in source domain (RUGD) are extremely rare in the target domain (RELLIS), hence no prediction for those classes occurs especially considering the domain shift.}

\begin{table}
\caption{\small Quantitative comparison of different methods in UDA of RUGD$\rightarrow$RELLIS. mIoU* is the average mIoU over all of classes.}
\centering 
{
\begin{tabularx}{0.92\linewidth}{cccc|cc}
\hline\hline 
  Class & SO & DA & CA & CALI & ICALI \\[0.5ex]
\hline 
Dirt & 0.00 & 0.53 & \textbf{3.23} & 0.01 & 0.73 \\ [1ex]
Grass & 64.78 & 61.63 & 65.35 & 67.08 & \textbf{71.2}\\ [1ex]
Tree & 40.79 & 45.93 & 41.51 & \textbf{55.80} & 54.32 \\ [1ex]
Sky & 45.07 & 67.00 & 2.31 & 72.99 & \textbf{75.64}\\ [1ex]
Building & 10.90 & \textbf{12.29} & 10.91 &10.28 & 10.02\\ [1ex]
mIoU* & 32.31 & 37.48 & 24.66 & 41.23 & \textbf{42.38}\\
\hline\hline 
\end{tabularx}
}
\label{tb:outdoor_quantitative} 
\end{table}
We observe similar trends as that in Table~\ref{tb:city_quantitative}. More specifically, both tables show that CA has the negative transfer (compared with SO) issue for either sim2real or real2real UDA. However, if we constrain the training of CA with DA, as in our proposed model CALI, the performance will be remarkably improved. Some qualitative results are shown in Fig.~\ref{fig:outdoor_images}. However, if we compare CALI and ICALI, the gain for the setting of RUGD$\rightarrow$RELLIS is much less significant than the one in Table \ref{tb:city_quantitative}. If we look at qualitative results in Fig. \ref{fig:outdoor_images}, some predictions of ICALI look even worse than CALI, e.g., the last two rows. This is because the mixture step in ICALI highlights under-performing classes only in the source domain (as we can only have reliable identification of well/under-performing classes using provided labels in the source domain). This works well when the label shift between the source domain and the target domain is mild, e.g., the adaptation of GTA5$\rightarrow$Cityscapes. However, in the adaptation of RUGD$\rightarrow$RELLIS, the label shift is remarkable --- the proportion of classes has significantly changed and only a few of the classes in the source domain appear in the target domain. In this case, highlighting classes that rarely appear in the target domain might cause misunderstanding and demolish the performance.

\subsubsection{RUGD$\rightarrow$MESH} 
Our MESH dataset contains only unlabeled images that restrict us to show only a qualitative comparison for the UDA of RUGD$\rightarrow$MESH, as shown in Fig.~\ref{fig:mesh_images}. We have collected data in winter forest environments, which are significantly different than the images in the source domain (RUGD) - collected in a different season, e.g., summer or spring. These cross-season scenarios make the prediction more challenging. However, it is more practical to evaluate the UDA performance of cross-season scenarios, as we might have to deploy our robot at any time, even with extreme weather conditions, but our available datasets might be far from covering every season and every weather condition. From Fig.~\ref{fig:mesh_images}, we can still see the obvious advantages of our proposed CALI model over other baselines. Since the label shift between RUGD and MESH is still large, the advantage of ICALI over CALI is still not remarkable.

\begin{figure*} 
{
  \centering
    \subfigure[]
  	{\label{fig:city_comparison_1}\includegraphics[width=0.31\linewidth]{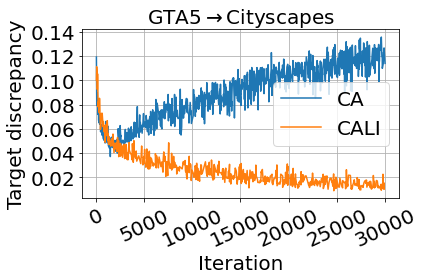}}
  	\subfigure[]
  	{\label{fig:city_comparison_2}\includegraphics[width=0.31\linewidth]{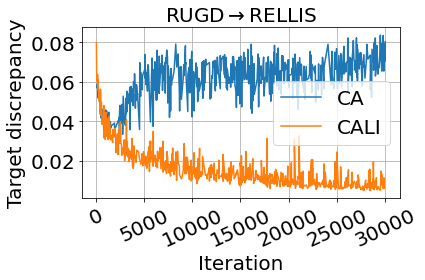}}  
  	\subfigure[]
  	{\label{fig:city_comparison_3}\includegraphics[width=0.31\linewidth]{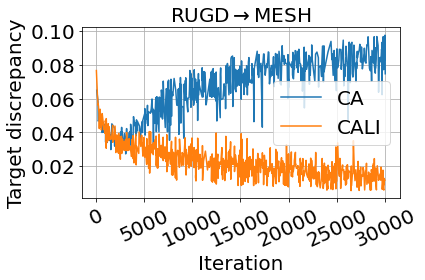}}  
  \caption{\small Target discrepancy changes during training process of (a) GTA5$\rightarrow$Cityscapes; (b) RUGD$\rightarrow$RELLIS; and (c) RUGD$\rightarrow$MESH.
  }
\label{fig:dis_comparison}  
}
\end{figure*}

\begin{figure} 
{
  \centering
    \subfigure[]
  	{\label{fig:d_outputs}\includegraphics[width=0.48\linewidth]{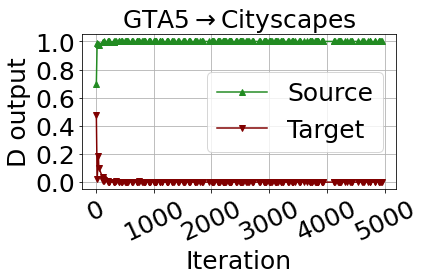}}
  	\subfigure[]
  	{\label{fig:minmax_dis}\includegraphics[width=0.48\linewidth]{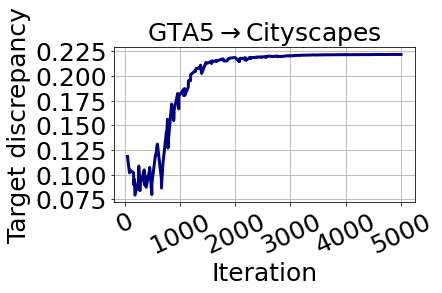}} 
  \caption{\small Using minmax can cause the collapse of training.
  }
\label{fig:minmax_curves}  
}
\end{figure}

\begin{figure}
{
\centering
  {\includegraphics[width=\linewidth]{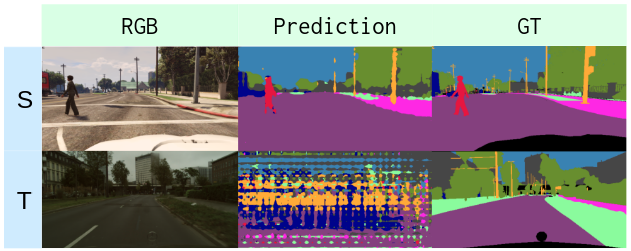} 
  } 
\caption{\small An example of collapsed trained model using minmax. }  \vspace{-10pt}
\label{fig:minmax_failure} 
}
\end{figure}

\begin{figure*}
{
\centering
  {\includegraphics[width=\linewidth]{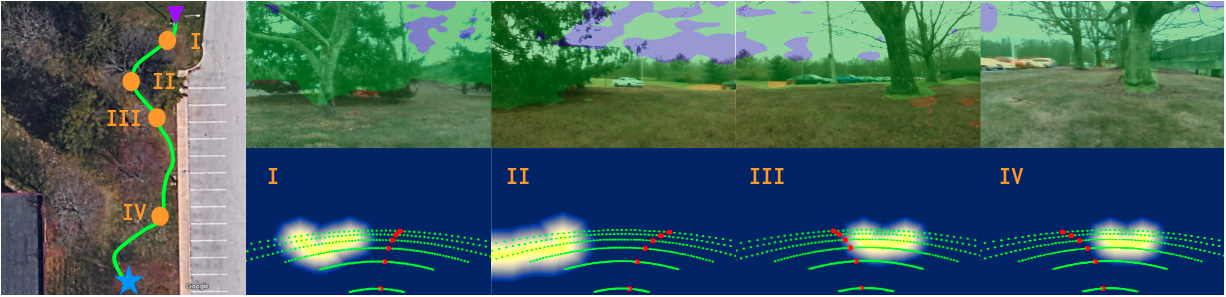} 
  } 
\caption{\small Navigation behaviors in MESH$\#1$ environment. The left-most column: top-down view of the environment; Purple triangle: the starting point; Blue star: the target point; We also show the segmentation (top row) and planning results (bottom row) at four different moments during the navigation, as shown from the second column to the last one. }  
\label{fig:nav_1} 
}
\end{figure*}

\begin{figure*}
{
\centering
  {\includegraphics[width=\linewidth]{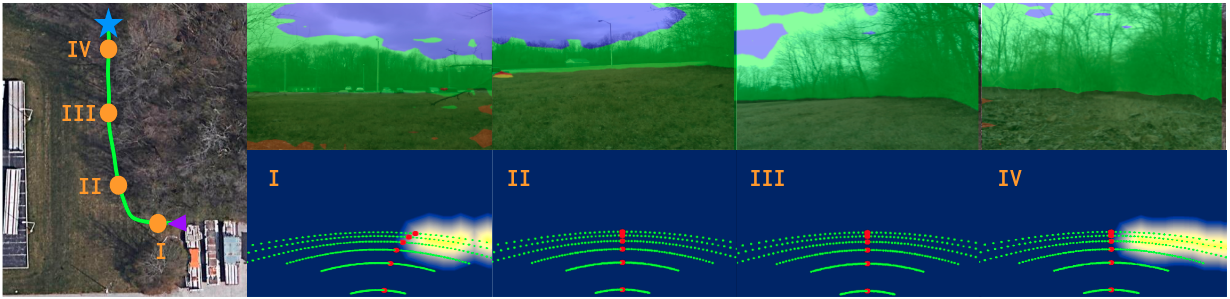} 
  } 
\caption{\small Navigation behaviors in MESH$\#2$ environment. Same legends with Fig.~\ref{fig:nav_1}.} 
\label{fig:nav_2} 
}
\end{figure*}

\subsection{Discussions}
\label{sec:ablation_studies}
In this section, we aim to discuss our model (CALI) behaviors in more details. Specifically, first we will explain the advantages of CALI over CA from the perspective of training process. Second, we will show the vital influence of mistakenly using the wrong order of adversarial training.

The most important part in CA is the discrepancy between the two classifiers, which is the only training force for the functionality of CA. It has been empirically studied in \citet{saito2018maximum} that the target prediction accuracy will increase as the target discrepancy is decreasing, hence the discrepancy is also an indicator showing if the training is on the right track. We compare the target discrepancy changes of CALI and our baseline CA in Fig.~\ref{fig:dis_comparison}, where the curves for the three UDA scenarios are presented from (a) to (c) and we only show the data before iteration 30k. It can be seen that before around iteration 2k, the target discrepancy of both CALI and CA are drastically decreasing, but thereafter, the discrepancy of CA starts to increase. On the other hand, if we impose a DA constraint over the same CA (iteratively), leading to our proposed CALI, then the target discrepancy will be decreasing as expected. This validates that integrating DA and CA will make the training process of CA more stable, thus improving the target prediction accuracy.

As mentioned in Algorithm 1,
we have to use adversarial training order of $\max_{\psi_D} \min_{\phi_G}$, instead of using  $\min_{\phi_G} \max_{\psi_D}$. The reason for this is related to our designed net structure. Following the guidance of Eq.~(\ref{eq:bound_relation}), we use the same input to the two classifiers and the domain discriminator, hence the discriminator has to receive the intermediate-level feature as the input. If we use the order of $\min_{\phi_G} \max_{\psi_D}$ in CALI, then the outputs of the discriminator will be like Fig.~\ref{fig:d_outputs}, where the domain discriminator of CALI will quickly converge to the optimal state and it can accurately discriminate if the feature is from source or target domain. In this case, the adversarial loss for updating the feature extractor will be near 0, hence the whole training fails, which is validated by changes of the target discrepancy curve, as shown in Fig.~\ref{fig:minmax_dis}, where the discrepancy value is decreasing in a small amount in the first few iterations and then quickly increase to a high level that shows the training is divergent and the model is collapsed. This is also verified by the prediction results at (and after) around iteration 1k, as shown in Fig.~\ref{fig:minmax_failure}, where the first row is the source images while the second row is the target images.

\subsection{Navigation Missions}
To further show the effectiveness of our proposed CALI model for real deployments,
we build a navigation system by combining the proposed CALI (trained with RUGD $\rightarrow$ MESH set-up) segmentation model with our visual planner. We test behaviors of our navigation system in two different forest environments (MESH$\#1$ in Fig.~\ref{fig:nav_1} and MESH$\#2$ in Fig.~\ref{fig:nav_2}), where our navigation system shows high reliability. 
In navigation tasks, the image resolution is $[400, 300]$, and the inference time for pure segmentation inference is around $33$ frame per second (FPS). However, since a complete perception system requires several post-processing steps, such as navigability definition, noise filtering, Scaled Euclidean Distance Field computation, motion primitive evaluation and so on, the response time for the whole perception pipeline (in python) is around $8$ FPS without any engineering optimization. The inference of segmentation for navigation is performed on an Nvidia Tesla T4 GPU. We set the linear velocity as $0.3m/s$ and control the angular velocity to track the selected motion primitive. The path length is $32.26m$ in Fig.~\ref{fig:nav_1} and $28.63m$ in Fig.~\ref{fig:nav_2}. Although the motion speed is slow in navigation tasks, as a proof of concept and with a very basic motion planner, the system behavior is as expected, and we have validated that the proposed CALI model is able to accomplish the navigation tasks in unstructured environments.

\section{Conclusion and Future Work} 
\label{sec:conclusion}
We present CALI, a novel unsupervised domain adaptation model specifically designed for semantic segmentation, which requires fine-grained alignments in the level of class features. We carefully investigate the relationship between a coarse alignment and a fine alignment in theory. The theoretical analysis guides the design of the model structure, losses, and training process. We have validated that the coarse alignment can serve as a constraint to the fine alignment and integrating the two alignments can boost the UDA performance for segmentation. The resultant model shows significant advantages over baselines in various challenging UDA scenarios, e.g., sim2real and real2real. 

To further improve this framework in the future, we observe one problem with our model during deployment --- segmentation boundaries can jump (vary fast) when the model makes highly-frequent predictions for images from the video captured by the onboard camera. Our model predicts segmentation for each frame independently and ignores the inter-frame coherence. This sometimes leads to a non-negligible disturbance to smooth navigation, as our visual planner highly relies on the segmentation boundaries to generate the SEDF for obstacle avoidance. To further increase the practicality of our model in real deployments, incorporating temporal correlation during training or inference is necessary to help maintain stable segmentation boundaries. In addition, we observe that the proposed ICALI model only has significant performance gain when the label shift between the source domain and the target domain is small. However, we usually have a large label shift between the source and the target in real deployments, e.g., RUGD$\rightarrow$RELLIS or RUGD$\rightarrow$MESH. Addressing this issue and generalizing ICALI to cases where large label shifts exist is worth more investigation in the future. 

\section{Appendix}
\subsection{Proof of \textbf{Theorem 2}}
\label{sec:proof}
For a hypothesis $h$,
\begin{equation}
    \label{eq:proof_theorem2}
    \begin{aligned}
    \epsilon_t(h) &\leq \epsilon_t(h^{*})+\epsilon_t(h, h^{*})\\
    &=\epsilon_t(h^{*})+\epsilon_s(h, h^{*})-\epsilon_s(h, h^{*})+\epsilon_t(h, h^{*})\\
    &\leq \epsilon_t(h^{*})+\epsilon_s(h, h^{*})+|\epsilon_t(h, h^{*}) - \epsilon_s(h, h^{*})|\\
    &\leq \epsilon_t(h^{*})+\epsilon_s(h, h^{*})+\frac{1}{2}d_{\mathcal{H}\Delta\mathcal{H}}(\mathcal{D}_s, \mathcal{D}_t)\\
    &\leq \epsilon_t(h^{*})+\epsilon_s(h)+\epsilon_s(h^{*})+\frac{1}{2}d_{\mathcal{H}\Delta\mathcal{H}}(\mathcal{D}_s, \mathcal{D}_t)\\
    &= \epsilon_s(h)+\frac{1}{2}d_{\mathcal{H}\Delta\mathcal{H}}(\mathcal{D}_s, \mathcal{D}_t)+\epsilon_s(h^{*})+\epsilon_t(h^{*})\\
    &= \epsilon_s(h)+\frac{1}{2}d_{\mathcal{H}\Delta\mathcal{H}}(\mathcal{D}_s, \mathcal{D}_t)+\lambda\\
    &=\epsilon_s(h)+\sup_{h, h^{'}\in \mathcal{H}}|\text{P}_{\mathbf{x}\sim \mathcal{D}_s}\left [ h(\mathbf{x})\neq h^{'}(\mathbf{x}) \right ] -\\ &~~~~~~~~~~~~~~~~~~~~~~~\text{P}_{\mathbf{x}\sim \mathcal{D}_s}\left [ h(\mathbf{x})\neq h^{'}(\mathbf{x}) \right ]|+\lambda\\
    &=\epsilon_s(h)+\sup_{g\in \mathcal{H}\Delta\mathcal{H}}|\text{P}_{\mathbf{x}\sim \mathcal{D}_s}\left [ g(\mathbf{x})=1) \right ] -\\ &~~~~~~~~~~~~~~~~~~~~~~~\text{P}_{\mathbf{x}\sim \mathcal{D}_t}\left [ g(\mathbf{x})=1 \right ]|+\lambda\\
    &= \epsilon_s(h)+\sup_{g\in \mathcal{H}\Delta\mathcal{H}}| \text{P}_{\mathbf{x}\sim \mathcal{D}_s}\left [ g(\mathbf{x})=1) \right ] +\\ &~~~~~~~~~~~~~~~~~~~~~~~\text{P}_{\mathbf{x}\sim \mathcal{D}_t}\left [ g(\mathbf{x})=0 \right ]-1|+\lambda\\
    &\leq\epsilon_s(h)+\sup_{g\in \mathcal{H}\Delta\mathcal{H}}| \text{P}_{\mathbf{x}\sim \mathcal{D}_s}\left [ g(\mathbf{x})=1) \right ] +\\ &~~~~~~~~~~~~~~~\text{P}_{\mathbf{x}\sim \mathcal{D}_t}\left [ g(\mathbf{x})=0 \right ]| - \inf_{g\in \mathcal{H}\Delta\mathcal{H}} 1+\lambda\\
    &=\epsilon_s(h)+\sup_{g\in \mathcal{H}\Delta\mathcal{H}}| \text{P}_{\mathbf{x}\sim \mathcal{D}_s}\left [ g(\mathbf{x})=1) \right ] +\\ &~~~~~~~~~~~~~~~\text{P}_{\mathbf{x}\sim \mathcal{D}_t}\left [ g(\mathbf{x})=0 \right ]|+\lambda-1\\
    &=\epsilon_s(h)+\frac{1}{2}d_{\mathcal{H}}(\mathcal{D}_s, \mathcal{D}_t)+1+\lambda-1\\
    &=\epsilon_s(h)+\frac{1}{2}d_{\mathcal{H}}(\mathcal{D}_s, \mathcal{D}_t)+\lambda,
    \end{aligned}
\end{equation}
where $\lambda=\epsilon_s(h^{*})+\epsilon_t(h^{*})$ and $h^{*}$ is the ideal joint hypothesis (see the \textbf{Definition 2} in Section 4.2 of \citet{ben2010theory}).

We have the $4^{th}$, and the $8^{th}$ line because of the \textbf{Lemma 3} \citep{ben2010theory}; the $5^{th}$ line because of the \textbf{Theorem 2} \citep{ben2010theory}; the last second line because of the \textbf{Lemma 2} \citep{ben2010theory}. We have the $11^{th}$ line because $\sup|f_1 - f_2|= \sup f_1-\inf f_2 \leq \sup |f_1| - \inf f_2.~~~~~~~~~~~~~~~~~~~~~~~~~~~~~~~~~~~~~~~~~~\blacksquare$\\

\subsection{Remapping of Label Space}
\label{sec:remapping}
\begin{figure}[t]
{
\centering
  {\includegraphics[width=\linewidth]{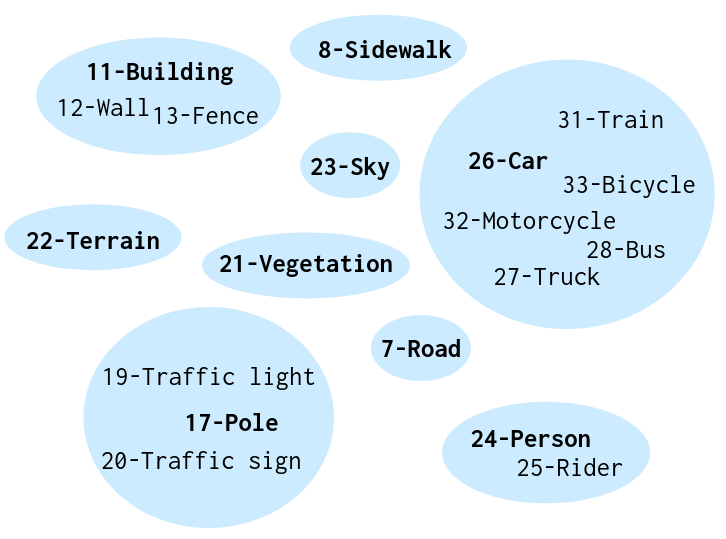} 
  } 
\caption{Lable remapping for GTA5$\rightarrow$Cityscapes. Name of each new group is marked as bold.}
\label{fig:city_cloud} 
}
\end{figure}

\begin{figure}
{
\centering
  {\includegraphics[width=\linewidth]{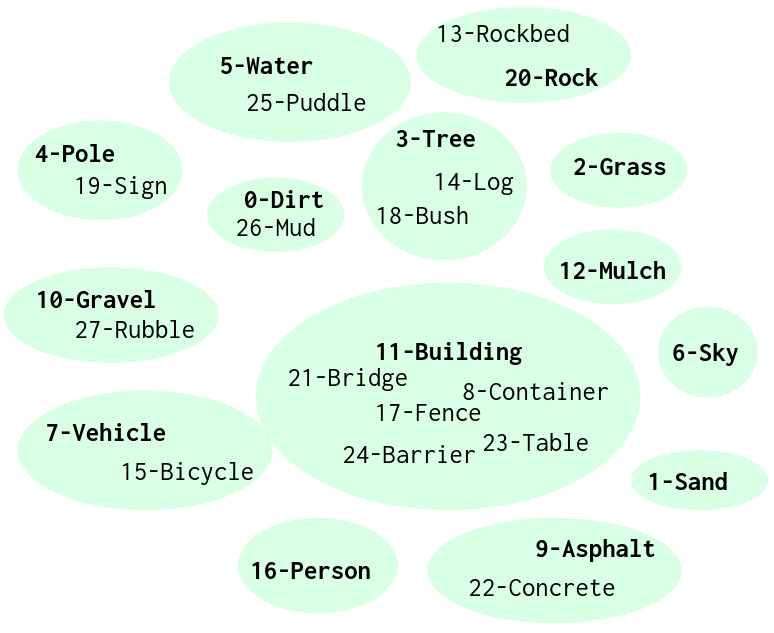} 
  } 
\caption{Lable remapping for RUGD$\rightarrow$RELLIS and RUGD$\rightarrow$MESH. Name of each new group is marked as bold.}  \vspace{-10pt}
\label{fig:outdoor_cloud} 
}
\end{figure}

We regroup the original label classes according to the semantic similarities among classes. In GTA5 and City-scapes, we cluster the building, wall and fence as the same category; traffic light, traffic sign and pole as the same group; car, train. bicycle, motorcycle, bus and truck as the same class; and treat the person and rider as the same one. See Fig.~\ref{fig:city_cloud}. Similarly, we also have regroupings for classes in RUGD and RELLIS, as can be seen in Fig.~\ref{fig:outdoor_cloud}.

\section{Statements and Declarations}

\subsection*{\textbf{Acknowledgements and Funding}} \vspace{-10pt}
We acknowledge the supports of NSF with grant number 2006886 as well as ARL with grant number W911NF-20-2-0099. We are also grateful for the computational resources provided by the Amazon AWS Machine Learning Research Award.

\subsection*{\textbf{Authors Contributions}} \vspace{-10pt}
Zheng Chen contributed to the algorithm design and the theoretical analysis.
The experiment validation was performed by Zheng Chen and Durgakant Pushp. 
Jason Gregory provided fruitful discussions on the design of ICALI. 
The manuscript was written by Zheng Chen and all authors commented and revised the manuscript. 
Lantao Liu contributed to the top design and general guidance of this work. All authors read and approved the final manuscript.

\subsection*{\textbf{Competing Interests}} \vspace{-10pt}
The authors have no relevant financial or non-financial interests to disclose.

\subsection*{\textbf{Ethics Approval}} \vspace{-10pt}
This paper does not involve human or animal subjects. No ethical approval is
required.

\subsection*{\textbf{Consent For Participation}} \vspace{-10pt}
This paper does not involve human subjects to participate in the study.

\subsection*{\textbf{Consent For Publication}} \vspace{-10pt}
The paper does not contain any data/figures collected from human participants. No consent is required to publish the manuscript

{
\bibliographystyle{plainnat}
\bibliography{reference.bib}
}

\end{document}